\documentclass[final]{cvpr}

\usepackage{times}
\usepackage{epsfig}
\usepackage{graphicx}
\usepackage{amsmath}
\usepackage{amssymb}
\usepackage{float}
\usepackage{booktabs}
\usepackage{multirow}
\usepackage{gensymb}
\usepackage{placeins}
\usepackage{paralist}
\usepackage{pifont}
\usepackage{caption}
\usepackage{subcaption}
\usepackage[dvipsnames]{xcolor}
\usepackage{color}
\usepackage{xr-hyper}

\usepackage{caption}
\usepackage{hyperref}
\hypersetup{pagebackref=true,breaklinks=true,colorlinks,bookmarks=false}

\newcommand{\markgood}[1]{{\color{ForestGreen}#1}}

\newcommand{\tf}[1]{\mathbf{#1}}
\newcommand{\R}{\mathbb{R}}

\newcommand{\cmark}{\ding{51}}
\newcommand{\xmark}{\ding{55}}

\makeatletter
\newcommand*{\addFileDependency}[1]{% argument=file name and extension
  \typeout{(#1)}
  \@addtofilelist{#1}
  \IfFileExists{#1}{}{\typeout{No file #1.}}
}
\makeatother

\def\cvprPaperID{1291} % *** Enter the CVPR Paper ID here
\def\confYear{CVPR 2021}

\begin{document}

\nocite{PIFu}  % Figure out how to talk about these!
\nocite{DISN}

%%%%%%%%% TITLE
\title{pixelNeRF: Neural Radiance Fields from One or Few Images}

% NO PAGE NUMBER?
\pagenumbering{gobble}

\author{
Alex Yu \qquad 
Vickie Ye \qquad
Matthew Tancik\qquad 
Angjoo Kanazawa\\
% INSTITUTION
UC Berkeley\\
% EMAIL
% {\tt\small
%     \{sxyu, vye, tancik, kanazawa\}@berkeley.edu
}

% \makeatletter
% \g@addto@macro\@maketitle{
%   \begin{figure}[H]
%   \setlength{\linewidth}{\textwidth}
%   \setlength{\hsize}{\textwidth}
%   \centering
% \begin{center}
%     %\fbox{\rule{0pt}{2.0in} \rule{0.95\linewidth}{0pt}}
%     \includegraphics[width=\linewidth]{figures/teaser_v0.pdf}
% \end{center}
%   \caption{Comparison to 3-view NeRF on the DTU dataset.}
  
% \label{fig:teaser}
%   \end{figure}
%   \vspace*{8pt}
% }
% \makeatother
% \maketitle

%%%%%%%%% ABSTRACT
\twocolumn[{%
  \renewcommand\twocolumn[1][]{#1}%
\maketitle
\thispagestyle{empty}
\begin{center}
  \newcommand{\teaserwidth}{0.9\textwidth}
  \vspace{-0.2in}
  %\vspace{0.01in}
  \centerline{
    \includegraphics[width=\teaserwidth]{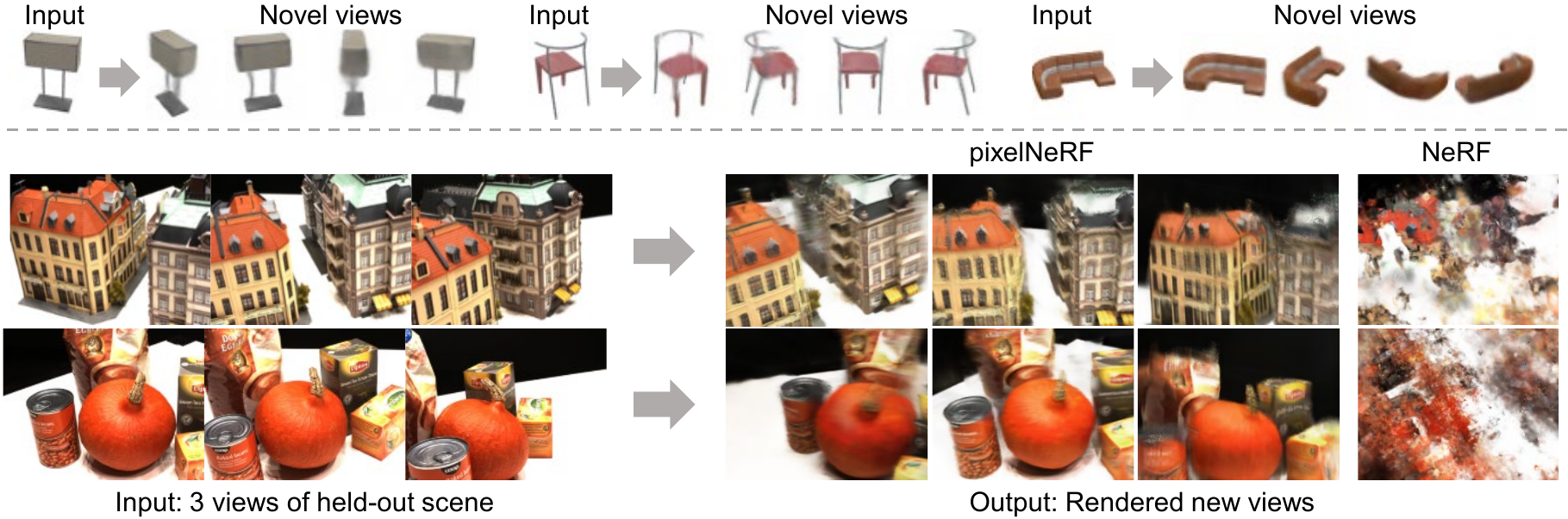}
    }
    \captionof{figure}{\textbf{NeRF from one or few images.} We present pixelNeRF, a learning framework that predicts a Neural Radiance Field (NeRF) representation from a single (top) or few posed images (bottom).
      %PixelNeRF can be trained on a dataset where it learns a 3D scene prior that allows it to generate plausible views from very few input images.
      %While NeRF is an optimization based framework that is fit to each scene independently, 
      %While original NeRF optimizes NeRF to each scene independently, 
      PixelNeRF can be trained on a set of multi-view images, allowing it to generate plausible novel view synthesis from very few input images without test-time optimization (bottom left). In contrast, NeRF has no generalization capabilities and performs poorly when only three input views are available (bottom right).
      }
  \vspace{-0.1in}
  %\vspace{-1em}
  \label{fig:teaser}
 \end{center}%
}]
\begin{abstract}
%\vspace{-5em}
We propose pixelNeRF, a learning framework that predicts a continuous neural scene representation conditioned on
one or few input images.
The existing approach for
constructing neural radiance fields \cite{NeRF}
involves optimizing the representation to every scene independently, requiring many calibrated views and significant compute time. 
We take a step towards resolving these shortcomings
by introducing an architecture that conditions a NeRF on image inputs in a fully convolutional manner. This allows the network to be trained across multiple scenes to learn a scene prior, enabling it to perform novel view synthesis in a feed-forward manner from a sparse set of views (as few as one).
Leveraging the volume rendering approach of NeRF, our model can be trained directly from images with no explicit 3D supervision.
We conduct extensive experiments on ShapeNet benchmarks for single image novel view synthesis tasks with held-out objects as well as entire unseen categories. We further demonstrate the flexibility of pixelNeRF by demonstrating it on multi-object ShapeNet scenes and real scenes from the DTU dataset. In all cases, pixelNeRF outperforms current state-of-the-art baselines for novel view synthesis and single image 3D reconstruction. 
For the video and code, please visit
the project website: 
\url{https://alexyu.net/pixelnerf}.
\pagebreak[3]
\end{abstract}

\vspace{-1em}
\section{Introduction}
%% Motivation/problem setup
We study the problem of synthesizing novel views of a scene from a sparse set of input views.
% The dream is to, given a few input images, be able to quickly reconstruct new photorealistic views of the same scene.
% Why do I care about you solving it? Does it make a difference in my life?
% How do people solve it normally, and what are the shortcomings of these methods?
% What are we doing? Why will we succeed?
This long-standing problem has recently seen progress due to advances in differentiable neural rendering~\cite{NeRF, NSVF, NeRFW, SRN}.
Across these approaches, a 3D scene is represented with a neural network, which can then be rendered into 2D views.
%Such representations allow for both implicit and explicit control of scene properties in new views. 
%Notably, recently method of neural radiance fields (NeRF) \cite{NeRF} accurately represent a scene by implicitly encoding volumetric density and color within a neural network.
Notably, the recent method neural radiance fields (NeRF)~\cite{NeRF} has shown impressive performance on novel view synthesis of a specific scene by implicitly encoding volumetric density and color through a neural network. While NeRF can render photorealistic novel views, it is often impractical as it requires a large number of posed images and a lengthy per-scene optimization.

%% AK: What we do
In this paper, we address these shortcomings by proposing pixelNeRF, a learning framework that enables predicting NeRFs from one or several images in a feed-forward manner. Unlike the original NeRF network, which does not make use of any image features, pixelNeRF takes spatial image features aligned to each pixel as an input. This image conditioning allows the framework to be trained on a set of multi-view images, where it can learn scene priors to perform view synthesis from one or few input views. In contrast, NeRF is unable to generalize and performs poorly when few input images are available, as shown in Fig.~\ref{fig:teaser}. 
%as one of the input to the MLP that outputs density and color ouptut at each query spatial location. 

%% How we do
Specifically, we condition NeRF on input images by first computing a fully convolutional image feature grid from the input image. Then for each query spatial point $\tf x$ and viewing direction $\tf d$ of interest in the view coordinate frame, we sample the corresponding image feature via projection and bilinear interpolation. The query specification is sent along with the image features to the \textit{NeRF network} that outputs density and color, where the spatial image features are fed to each layer as a residual. When more than one image is available, the inputs are first encoded into a latent representation in each camera's coordinate frame, which are then pooled in an intermediate layer prior to predicting the color and density. The model is supervised with a reconstruction loss between a ground truth image and a view rendered using conventional volume rendering techniques.
This framework is illustrated in Fig.~\ref{fig:arch}.

%% Good things
PixelNeRF has many desirable properties for few-view novel-view synthesis. First, pixelNeRF can be trained on a dataset of multi-view images without additional supervision such as ground truth 3D shape or object masks. Second, pixelNeRF predicts a NeRF representation in the camera coordinate system of the input image instead of a canonical coordinate frame.
%, which is integral for generalization to  
This is not only integral for generalization to unseen scenes and object categories~\cite{What3d, Shin2018}, but also for flexibility, since no clear canonical coordinate system exists on scenes with multiple objects or real scenes.  
%Unlike prior work that operate on canonical frame, this design allows pixelNeRF to be flexible to the type of input scene allowing it to train on scenes with multiple objects or arbitrary scenes. 
Third, it is fully convolutional, allowing it to preserve the spatial alignment between the image and the output 3D representation. %Fourth, while pixelNeRF can be further optimized at test time, pixelNeRF is able to output a plausible novel-view synthesis in a feed-forward manner. 
Lastly, pixelNeRF can incorporate a variable number of posed input views at test time without requiring any test-time optimization.

We conduct an extensive series of experiments on synthetic and real image datasets to evaluate the efficacy of our framework, going beyond the usual set of ShapeNet experiments %for novel view synthesis
to demonstrate its flexibility. Our experiments show that pixelNeRF can generate novel views from a single image input for both category-specific and category-agnostic settings, even in the case of unseen object categories.
Further, we test the flexibility of our framework, both with a new multi-object benchmark for ShapeNet, where pixelNeRF outperforms prior approaches,
and with simulation-to-real transfer demonstration on real car images.
Lastly, we test capabilities of pixelNeRF on real images using the DTU dataset~\cite{DTU},
where despite being trained on under 100 scenes, it can generate plausible novel views of a real scene from three posed input views.

\begin{table}
    %\begin{center}
    \centering
    \setlength{\tabcolsep}{1pt}

\begin{tabular}{@{}ccccccc@{}}
\toprule
                    & NeRF           & DISN             & ONet           & DVR            & SRN            & Ours             \\ \midrule
Learns scene prior? & \xmark             & \markgood{\cmark}   & \markgood{\cmark} & \markgood{\cmark} & \markgood{\cmark} & \markgood{\cmark}   \\
Supervision         & \markgood{2D}  & 3D               & 3D             & \markgood{2D}  & \markgood{2D}  & \markgood{2D}    \\
Image features      & \xmark           & \markgood{Local} & Global         & Global         & \xmark           & \markgood{Local} \\
Allows multi-view?  & \markgood{\cmark} & \markgood{\cmark}   & \xmark             & \xmark             & \markgood{\cmark} & \markgood{\cmark}   \\
View space?    & -              & \xmark              & \xmark            & \xmark            & \xmark            & \markgood{\cmark}    \\ \bottomrule
\end{tabular}
    %\end{center}
\caption{
    \textbf{A comparison with prior works reconstructing 
    neural scene representations.}
    % I settled with neural scene repr. instead of implicit
    The proposed approach learns a scene prior for one or few-view reconstruction
    using only multi-view 2D image supervision. 
    Unlike previous methods in this regime,
    we do not require a consistent canonical space across
    the training corpus.
    Moreover, we incorporate local image features to preserve local information which is
    in contrast to methods that compress the structure and appearance 
    into a single latent vector such as Occupancy Networks (ONet)~\cite{OccupancyNetworks} and DVR~\cite{DVR}.
}
    \label{tab:related_work_comparison}
    \vspace{-1em}
\end{table}

\section{Related Work}
\label{sec:rel_work}

\noindent\textbf{Novel View Synthesis.}
The long-standing problem of novel view synthesis entails constructing new views of a scene from a set of input views.
Early work achieved photorealistic results but required densely captured views of the scene~\cite{levoy1996light, gortler1996lumigraph}.
Recent work has made rapid progress toward photorealism for both wider ranges of novel views and sparser sets of input views, by using 3D representations based on neural networks~\cite{NeRF, NeuralVolumes, meshry2019neural, DeepVoxels, thies2019deferred, dai2020neural}.
However, because these approaches fit a single model to each scene, they require many input views and substantial optimization time per scene.

% Other recent learning-based approaches use representations conditioned on images.
% There are lots of novel view synthesis: zhou2017view
% Recent work consider predicting this from single images: tucker2020singleview, Sih3DP20
% But these rely on 2.5D representations and therefore limited in the range of..
There are methods that can predict novel view from few input views or even single images by learning shared priors across scenes. 
Methods in the tradition of~\cite{shade1998layered, buehler2001unstructured} use depth-guided image interpolation~\cite{zhou2017view, DeepStereo, riegler2020free}.
More recently, the problem of predicting novel views from a single image has been explored~\cite{tucker2020singleview, wiles2020synsin, Shih3DP20,chen2019monocular}. However, these methods employ 2.5D representations, and are therefore limited in the range of camera motions they can synthesize. In this work we infer a 3D volumetric NeRF representation, which allows novel view synthesis from larger baselines.

Sitzmann et al.~\cite{SRN} introduces a representation based on a continuous 3D feature space
% that can be either fit to a single scene or used to 
to learn a prior across scene instances.
However, using the learned prior at test time requires further optimization with known absolute camera poses.
In contrast, our approach is completely feed-forward and only requires relative camera poses.
We offer extensive comparisons with this approach to demonstrate the advantages our design affords.
Lastly, note that concurrent work \cite{GRF} adds image features to NeRF. 
A key difference is that we operate in view rather than canonical space, which makes our approach applicable in more general settings. 
Moreover, we extensively demonstrate our method's performance in few-shot view synthesis, while GRF shows very limited quantitative results for this task.

\vspace{0.1in}
\noindent\textbf{Learning-based 3D reconstruction.}
Advances in deep learning have led to rapid progress in single-view or multi-view 3D reconstruction.
Many approaches~\cite{kar2017learning, AtlasNet, Pixel2Mesh, GenRe, DeepVoxels, PIFu, DISN, OccupancyNetworks, CoReNet} propose learning frameworks with various 3D representations that require ground-truth 3D models for supervision.
%Single view 3D reconstruction has largely focused on simple, object-centric scenes.
%Several single view reconstruction methods \cite{AtlasNet, Pixel2Mesh, OccupancyNetworks,  GenRe, DeepVoxels, PIFu, DISN, CoReNet} require ground truth 3D model supervision.
Multi-view supervision \cite{PTN, DRC, SoftRas, Liu2019, SRN, DVR, ENR, Bautista_2021_WACV} is less restrictive and more ecologically plausible. However, many of these methods~\cite{PTN, DRC, SoftRas, Liu2019, DVR} require object masks; in contrast,
pixelNeRF can be trained from images alone,
allowing it to be applied to 
scenes of two objects without modification.
%more readily available, and is shown to offer better performance to new datasets \cite{godard2017unsupervised}.
% \ak{Add something about image inputs. Most of these approaches condition their models on image inputs, many use global features, which cannot preserve local features, actually bring that 3rd paragraph content here. Just two paragraphs for RW is fine.}

Most single-view 3D reconstruction methods condition neural 3D representations on input images.
The majority employs global image features~\cite{park2019deepsdf, chen2019learning, DVR, OccupancyNetworks, ENR}, which, while memory efficient, cannot preserve details that are present in the image and often lead to retrieval-like results. %, since it is implausible that the appearance of complex scenes can be compressed into a short latent vector. 
Spatially-aligned \textit{local} image features have been shown to achieve detailed reconstructions from a single view~\cite{DISN, PIFu}.
However, both of these methods require 3D supervision. 
Our method is inspired by these approaches, but only requires multi-view supervision. % by way of the NeRF representation.

Within existing methods, the types of scenes that can be reconstructed are limited, particularly so for object-centric approaches (e.g.~\cite{Pixel2Mesh, SoftRas, AtlasNet, DRC, DeepVoxels, GenRe, OccupancyNetworks, DISN, DVR}).
CoReNet~\cite{CoReNet} % it is CoReNet not CoreNet
reconstructs scenes with multiple objects via a voxel grid with offsets, but it requires 3D supervision  including the identity and placement of objects.
In comparison, we formulate a scene-level learning framework that can in principle be trained to scenes of arbitrary structure.
% The recent method CoReNet~\cite{CoReNet} can represent multiple objects using a voxel grid with offsets but requires 3D supervision.
%In order to handle natural images ``in-the-wild", we aim to move beyond single-object representations and into scene representations.

\vspace{0.5em}
\noindent\textbf{Viewer-centric 3D reconstruction}
For the 3D learning task, prediction can be done either in a viewer-centered coordinate system, i.e. \textit{view space}, or in an object-centered coordinate system, i.e. \textit{canonical space}.
Most existing methods~\cite{DISN, OccupancyNetworks, DVR, SRN} predict in canonical space, where all objects of a semantic category are aligned to a consistent orientation.
While this makes learning spatial regularities easier,
using a canonical space 
inhibits prediction performance on unseen object categories and scenes with more than one object, where there is no pre-defined or well-defined canonical pose.
PixelNeRF operates in view-space, which 
%Moreover, training a model in view space 
has been shown to allow better reconstruction of unseen object categories in~\cite{Shin2018, Bautista_2021_WACV},
and discourages the memorization of the training set~\cite{What3d}.
%By designing our method to predict in view space, we enable our feed-forward model to better reconstruct previously unseen scenes or objects.
We summarize key aspects of our approach relative to prior work in Table~\ref{tab:related_work_comparison}. 

\section{Background: NeRF}
\label{sec:background}
%\ak{Similar to Nerf-W, explain, and then end with limitations, which we address.}
% Fundamentally, we aim to reconstruct a rich 3D representation from very few views of a scene, enabling novel view synthesis.
% In order to accelerate reconstruction
% and reduce the number of views required, we must be able to learn priors of 3D structure from image appearance.

We first briefly review the NeRF representation~\cite{NeRF}.
A NeRF encodes a scene as a continuous volumetric radiance field $f$ of color and density.
Specifically, for a 3D point $\tf x \in \R^3$ and viewing direction unit vector $\tf d \in \R^3$, $f$ returns a differential density $\sigma$ and RGB color $\tf c$:~\( f(\tf x, \tf d) = (\sigma, \tf c)\).
    
%\begin{equation}
%    \label{eq:nerf_func}
%\end{equation}
% Where $\tf c = (R, G, B) \in [0,1]^3$ denotes the color and $\sigma \in [0, \infty)$ denotes the differential density. 
The volumetric radiance field can then be rendered into a 2D image via
\begin{equation}
 %\begin{split}
     \hat{\mathbf{C}}(\mathbf{r}) = \int_{t_n}^{t_f} T(t) \sigma(t) \mathbf{c}(t) dt %\\
 %\end{split}
 \label{eq:rendering}
\end{equation}
where~$T(t) = \exp\big(- \int_{t_n}^{t} \sigma(s) \,ds \big)$ handles occlusion.
For a target view with pose $\tf P$, a camera ray can be parameterized as $\tf r(t) = \tf o + t \tf d$, with the ray origin (camera center) $\tf o \in \R^3$ and ray unit direction vector $\tf d \in \R^3$.
The integral is computed along $\tf r$ between pre-defined depth bounds $[t_n, t_f]$.
In practice, this integral is approximated with numerical quadrature by sampling points along each pixel ray.
% To render the implicit radiance field, NeRF uses the volumetric rendering integral in \eqref{eq:rendering}, 
% in which the radiance and density are integrated along a ray within pre-defined depth bounds $[t_n, t_f]$:

The rendered pixel value for camera ray $\tf r$ can then be compared against the corresponding ground truth pixel value, $\tf C(\tf r)$, for all the camera rays of the target view with pose $\tf P$.
The NeRF rendering loss is thus given by
% For a particular camera ray $\tf r$, let $\tf C (\tf r)$ denote the ground-truth color of the associated pixel.
\begin{equation}
    \mathcal{L} = 
    \sum_{\tf r \in \mathcal{R}(\tf P)}
        \left\lVert
   \hat{\tf C}(\tf r)
   -
   \tf C(\tf r)
    \right\rVert_2^2
    \label{eq:mseloss}
\end{equation}
where $\mathcal{R}(\tf P)$ is the set of all camera rays of target pose $\tf P$.

\vspace{.1in}
\noindent\textbf{Limitations} While NeRF achieves state of the art novel view synthesis results, it is an optimization-based approach using geometric consistency as the sole signal, similar to classical multiview stereo methods~\cite{RomeInDay, COLMAPMVS}.
As such each scene must be optimized individually, with no knowledge shared between scenes.
Not only is this time-consuming, but in the limit of single or extremely sparse views, it is unable to make use of any prior knowledge of the world to accelerate reconstruction or for shape completion. 

% It is worth noting that classical multi-view stereo methods such as have long been able to extract detailed scene geometry from multiple views.
% By taking an analogous optimization-only approach, NeRF shares the weakness of these
% methods in that 
\section{Image-conditioned NeRF}

\begin{figure*}[t]
\begin{center}
    \includegraphics[width=0.9\textwidth]{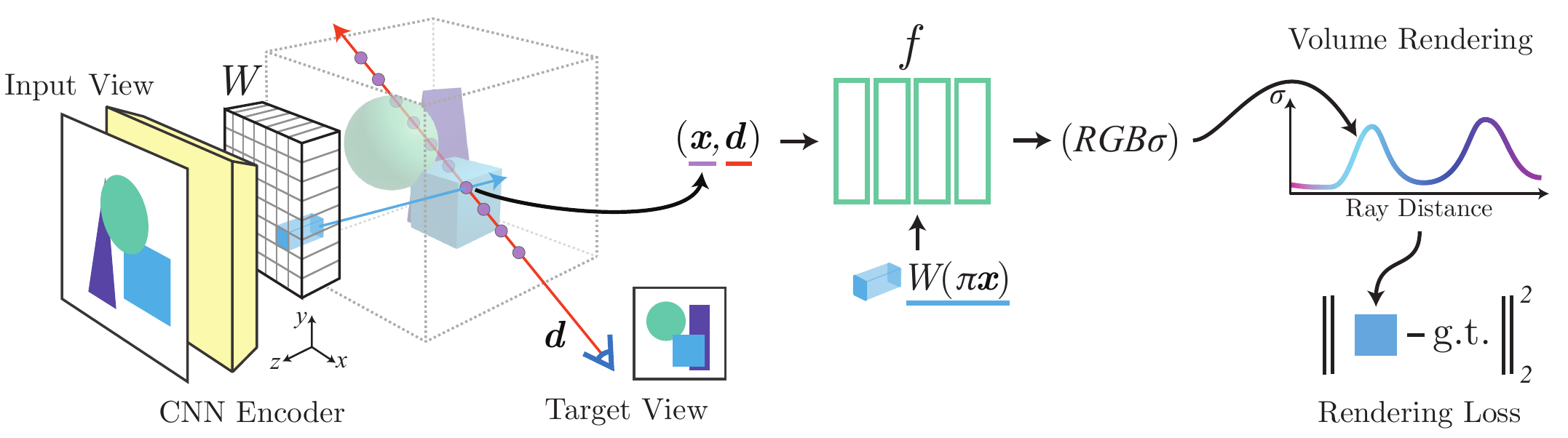}
\end{center}
\vspace{-6mm}
   \caption{\textbf{Proposed architecture in the single-view case.}
   For a query point $\tf x$ along a target camera ray with view direction $\tf d$,
   a corresponding image feature is extracted from the feature volume $\tf W$
   via projection and interpolation.
   This feature is then passed into the NeRF network $f$ along with the spatial coordinates.
   The output RGB and density value is volume-rendered and compared with the target pixel value.
   The coordinates $\mathbf{x}$ and $\mathbf{d}$ are in the camera coordinate system of the input view.
   }
   \label{fig:arch}
\vspace{-5mm}
\end{figure*}

%\ak{Opening paragraph on overview of our model: What our things can do/We propose an architecture that allows NeRF to be conditioned on an image. That is fully convolutional. Our approach is flexible to the number of input views, and allow inference with one or (explain what happens at training and what happens test time). We will first describe the single-view case, coord system, then multiview. Our approach can be trained without any gt 3D supervision. See Figure 2 for illustrations.  }
%Design choices when it comes to conditioning NeRF on images: we wanted to it be fully convolutional, able to take in multiple view inputs.

% We propose using fully convolutional image features as a way to inject prior information into the NeRF representation.
% Given these limitations, we propose conditioning NeRF with image features.
% Pixel-wise image features have been shown to capture local information for improving single view 3D reconstruction tasks \cite{PIFu}.
% allowing reconstruction of local details in single-view 3D reconstruction tasks \cite{PIFu}.
% capturing local information for 3D reconstruction tasks \cite{PIFu}.

To overcome the NeRF representation's inability to share knowledge between scenes, we propose an architecture to condition a NeRF on spatial image features.
Our model is comprised of two components:
a fully-convolutional image encoder $E$, which encodes the input image into a pixel-aligned feature grid, and a NeRF network $f$ which outputs color and density, given a spatial location and its corresponding encoded feature.
% The input image is first encoded into a pixel-aligned feature grid.
% Then, given a query spatial location,
% its corresponding image feature is extracted via camera projection and interpolation on this feature grid.
% The image feature, query location, and view direction are passed into the NeRF network to predict the corresponding color and density.
We choose to model the spatial query in the input view's camera space, rather than a canonical space, for the reasons discussed in $\S$~\ref{sec:rel_work}.
We validate this design choice in our experiments on unseen object categories ($\S$~\ref{sec:beyond_shapenet}) and complex unseen scenes ($\S$~\ref{sec:dtu}).
The model is trained with the volume rendering method and loss described in $\S$~\ref{sec:background}.

In the following, we first present our model for the single view case.
We then show how this formulation can be easily extended to incorporate multiple input
images.

% \ak{Give a short overview of how this works, similar to Intro paragraph 3, and refer to overview so people know what to expect. basically first paragraph of current 4.1}

% \ak{These two following topics sounds like they belong in related work. In fact we have few sentences for local image features. Now I think the choice of coordinate systmes should be something in the related section bc it's common for both single view 3D and view synthesis methods. Move them there and in this section only focus on describing how we do it. Instead of explaining that we do view and how its important, now that it's already discussed in background say how we actually do this. A simple sentence: While one way is to model the spatial query in canonical, we do it in camera view. }

\subsection{Single-Image pixelNeRF}
We now describe our approach to render novel views from one input image.
% Our model consists of two components: a fully-convolutional image encoder $E$, and a NeRF network $f$ in the style of Equation~\ref{eq:nerf_func}.
We fix our coordinate system as the \textit{view space} of the input image and specify positions and camera rays in this coordinate system.

% We pass in to the NeRF network features extracted from the input image, along with the 3D location and viewing direction in the input view coordinates.
Given a input image $\tf I$ of a scene, we first extract a feature volume $\tf W = E(\tf I)$.
Then, for a point on a camera ray $\tf x$, we retrieve the corresponding image feature by projecting $\tf x$ onto the image plane to the image coordinates $\pi(\tf x)$ using known intrinsics, then bilinearly interpolating between the pixelwise features to extract the feature vector $\tf W(\pi(\tf x))$.
The image features are then passed into the NeRF network, along with the position and view direction (both in the input view coordinate system), as
\begin{equation}
    f(\gamma(\tf x), \tf d; \tf W(\pi(\tf x))) = (\sigma, \tf c)
    \label{eq:icnerf_1view}
\end{equation}
where $\gamma(\cdot)$ is a positional encoding on $\tf x$ with $6$
exponentially increasing frequencies introduced in the original NeRF \cite{NeRF}. %\ak{describe how the image feature is fed into MLP!!!}
The image feature is incorporated as a residual at each layer;
see $\S$~\ref{sec:impl_detail} for more information.
We show our pipeline schematically in Fig.~\ref{fig:arch}.

In the few-shot view synthesis task, the query view direction is a useful signal for
determining the importance of a particular image feature in
the NeRF network.
% While NeRF uses view directions solely for modeling complex lighting effects, we posit that 
% they can also be helpful as an additional signal of image feature importance in pixelNeRF.
If the query view direction is similar to the input view orientation, the model can rely more directly on the input; if it is dissimilar, the model must leverage the learned prior. 
% rely more heavily on the input view directly, and 
% input view information can be directly incorporated, while one further away 
% hints the model to rely more heavily on the learned prior. 
Moreover, in the multi-view case, view directions could 
serve as a signal for the relevance and positioning of different views.
For this reason, we input the view directions at the beginning of the NeRF network.
% We validate these hypothesized benefits in an ablation study (Table \ref{tab:chairs_ablations}).

\subsection{Incorporating Multiple Views}
\label{sec:multiview}
%\ak{Look at PIFu and Multi-view Stereo Machine paper}

% Our architecture naturally extends to allow for an \textit{arbitrary} number of input views at test time.
Multiple views provide additional information about the scene and resolve 3D geometric ambiguities inherent to the single-view case.
We extend our model to allow for an arbitrary number of views at test time,
% which is important since input views can be specified in an arbitrary order.
%Moreover, as each view is handled entirely symmetrically, arbitrary permutations of the input views have no effect on the output.
% We note that while many works enforce the multi-view geometric consistency during training, 
% methods such as ENR~\cite{ENR} and GenRe~\cite{GenRe} can only make use of a single view at test time.
which distinguishes our method from existing approaches that are designed to only use single input view at test time.~\cite{ENR, GenRe}
Moreover, our formulation is independent of the choice of world space and the order of input views.

% In contrast, our architecture (like~\cite{LSM, DISN, PIFu}) has a natural extension allowing for an \textit{arbitrary} number of views at test time, as we will describe below.
% Please also refer to (some diagram to be added maybe in supplemental) for a visual interpretation.
% We would therefore prefer a method which can make use of additional views of a scene, improving as more are made available.

In the case that we have multiple input views of the scene,
we assume only that the relative camera poses are known. 
For purposes of explanation,
an arbitrary world coordinate system can be fixed
 for the scene.
We denote the $i$th input image as $\tf I^{(i)}$ and its associated 
% pose in this input frame as $\tf P_i = \begin{bmatrix}\tf R_i & \tf t_i\end{bmatrix}$.
camera transform from the world space to its view space as $\tf P^{(i)} = \begin{bmatrix}\tf R^{(i)} & \tf t^{(i)} \end{bmatrix}$.

For a new target camera ray, we transform a query point $\tf x$, with view direction $\tf d$, into the coordinate system of each input view $i$ with the world to camera transform as
\begin{equation}
\tf x^{(i)} = \tf P^{(i)} \tf x, \quad \tf d^{(i)} = \tf R^{(i)} \tf d
\end{equation}
To obtain the output density and color,
we process the coordinates and corresponding features 
in each view coordinate frame independently
and aggregate across the views within the NeRF network.
For ease of explanation, we denote the initial layers of the NeRF network as $f_1$,
which process inputs in each input view space separately, and the final layers as $f_2$,
which process the aggregated views.

We encode each input image into feature volume $\tf W^{(i)} = E(\tf I^{(i)})$.
For the view-space point $\tf x^{(i)}$, we extract the corresponding image feature from the feature volume $\tf W^{(i)}$ at the projected image coordinate $\pi(\tf x^{(i)})$.
We then pass these inputs into $f_1$ to obtain intermediate vectors:
% We then pass each tuple $(\tf x^{(i)}, \tf d^{(i)}, \tf W^{(i)}(\pi(\tf x^{(i)}))$ 
% into $f_1$:
\begin{equation}
    \tf V^{(i)} = f_1\left(\gamma(\tf x^{(i)}), \tf d^{(i)} ;\, \tf W^{(i)}\big( \pi(\tf x^{(i)}) \big) \right).
\end{equation}
The intermediate $\tf V^{(i)}$ are then aggregated with the average pooling operator $\psi$
and passed into a the final layers, denoted as $f_2$,
to obtain the predicted density and color:
\begin{equation}
\label{eq:icnerf_2view}
(\sigma, \tf c) = f_2\left(
\psi\left(\tf V^{(1)}, \ldots, \tf V^{(n)}\right)
\right).
\end{equation}
% where $\psi$ is an aggregation function invariant to permutations of the inputs (e.g.\ average).
% Observe that by only using relative quantities $\tf x^{(i)}, \tf d^{(i)}$ in the formulation,
% the choice of the world space where $\tf x, \tf d$ were defined has been erased and the output is independent of it. 
In the single-view special case, this simplifies to Equation~\ref{eq:icnerf_1view} with $f = f_2 \circ f_1$,
by considering the view space as the world space.
An illustration is provided in the supplemental.

\section{Experiments}
% \ak{Overview: We do extensive, both in shapnet standard benchmarks \& real image experiments. Note that our formulation is general that for these different types of experiments we have fixed architecture}

We extensively demonstrate our approach in three experimental categories:
\begin{inparaenum}[1)]
\item existing ShapeNet~\cite{ShapeNet} benchmarks for category-specific and category-agnostic view synthesis,
\item ShapeNet scenes with unseen categories and multiple objects, both of which require geometric priors instead of recognition, 
as well as domain transfer to real car photos and
% \item ShapeNet objects settings to challenge the learned geometric priors, rather than recognition priors, and
\item real scenes from the DTU MVS dataset~\cite{DTU}.
\end{inparaenum}

% We conduct extensive experiments using synthetic renderings of ShapeNet~\cite{ShapeNet} objects, including applications to multiple-object scenes and real car photos,
% in addition to evaluationwith  on real-world scenes from DTU MVS~\cite{DTU}.
% We use the same architecture across all setups and experiments.
\vspace{0.5em}
\noindent\textbf{Baselines}
For ShapeNet benchmarks,
we compare quantitatively and qualitatively to SRN~\cite{SRN} and DVR~\cite{DVR},
the current state-of-the-art in few-shot novel-view synthesis and 2D-supervised single-view reconstruction respectively.
We use the 2D multiview-supervised variant of DVR.
In the category-agnostic setting ($\S$~\ref{sec:multi_cat}), we also include
grayscale rendering of SoftRas~\cite{SoftRas} results.~\footnote{
Color inference is not supported by the public SoftRas code.
}
In the experiments with multiple ShapeNet objects, we compare with SRN, which can also model entire scenes.

For the experiment on the DTU dataset, we compare to NeRF~\cite{NeRF} trained on sparse views.
Because NeRF is a test-time optimization method, we train a separate model for each scene in the test set.

\vspace{0.1in}
\noindent\textbf{Metrics}
\label{sec:metrics}
We report the standard image quality metrics PSNR and SSIM~\cite{SSIM} for all evaluations.
We also include LPIPS~\cite{LPIPS}, which more accurately reflects human perception, in all evaluations except in the category-specific setup ($\S$~\ref{sec:single_cat}).
In this setting, we exactly follow the protocol of SRN~\cite{SRN} to remain comparable to prior works \cite{Tatarchenko2015, WorrallGTB18, GQN, ENR, GRF}, for which source code is unavailable.
% except in the category-specific setup ($\S$~\ref{sec:single_cat}), 
% where we exactly follow the protocol of SRN~\cite{SRN} in order to remain comparable to 
% prior works \cite{Tatarchenko2015, WorrallGTB18, GQN, ENR, GRF} for which source code is unavailable.

\vspace{0.1in}
\label{sec:impl_detail}
\noindent\textbf{Implementation Details}
For the image encoder $E$,
to capture both local and global information effectively,
we extract a feature pyramid from the image.
We use a ResNet34 backbone pretrained on ImageNet for our experiments.
% While in principle our architecture is general enough to accomodate any image encoder,
% for simplicity
% we selected  a ResNet34 backbone pretrained on ImageNet for our experiments.
Features are extracted prior to the first $4$ pooling layers, upsampled using bilinear interpolation, 
and concatenated to form latent vectors of size $512$ aligned to each pixel.
%We initialize the encoder weights with pretrained ImageNet weights.

To incorporate a point's corresponding image feature into the NeRF network $f$, we choose a
ResNet architecture with a residual modulation rather than simply concatenating the feature vector with the point's position and view direction.
Specifically, we feed the encoded position and view direction through the network
and add the image feature as a residual at the beginning of each ResNet block.
We train an independent linear layer for each block residual, in a similar manner as AdaIn and SPADE~\cite{AdaIn, SPADE}, a method previously used with success in ~\cite{OccupancyNetworks, DVR}.
Please refer to the supplemental for additional details.
% \section{ShapeNet Benchmarks}
% \ak{Overview: We do extensive, both in shapnet standard benchmarks \& real image experiments.}
% \ak{Note that our formulation is general that for these different types of experiments we have fixed architecture}

%\ak{This first sentence misses 2 objects and real world, hint them.}
% We conduct extensive experiments using synthetic renderings of ShapeNet~\cite{ShapeNet} objects, including applications to multiple-object scenes and real car photos,
% in addition to evaluation on real-world scenes from DTU MVS~\cite{DTU}.
% Note we use the same architecture across all setups, without additional design or architecture additions for each experiment.
\begin{figure}
    \centering
    \includegraphics[width=\linewidth]{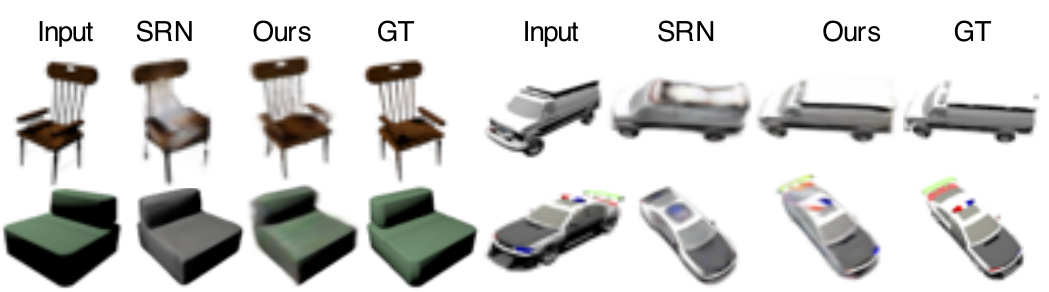}
    \caption{\textbf{Category-specific single-view reconstruction benchmark}.
    We train a separate model for cars and chairs and compare to SRN.
    The corresponding numbers may be found in Table~\ref{tab:single_cat}.
    }
    \label{fig:single_view_cars_chairs}
\end{figure}

\begin{figure}
    \centering
    \includegraphics[width=\linewidth]{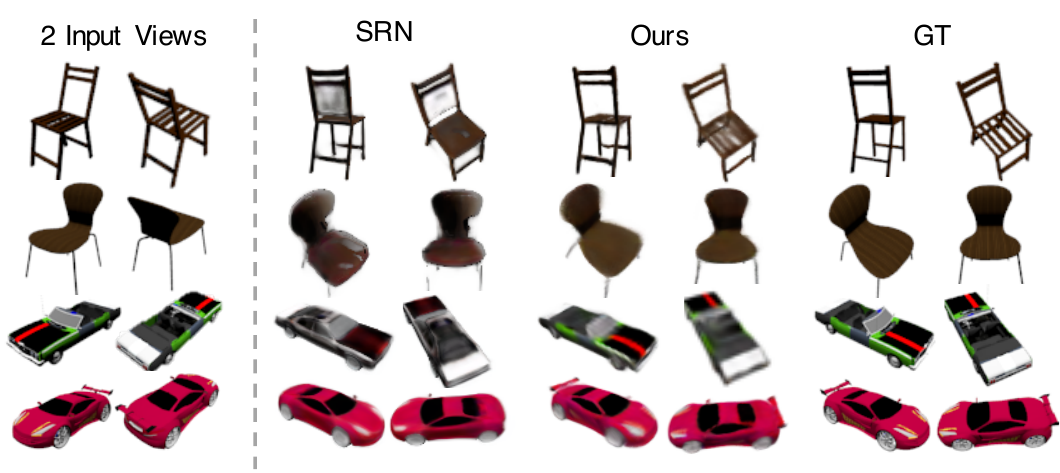}
    \caption{\textbf{Category-specific $2$-view reconstruction benchmark}.
    We provide two views (left) 
    to each model, and show two novel view renderings in each case (right).
    Please also refer to Table~\ref{tab:single_cat}.
    }
    \label{fig:single_view_cars_chairs}
\end{figure}

\subsection{ShapeNet Benchmarks}
We first evaluate our approach on category-specific and category-agnostic view synthesis tasks on ShapeNet.
% In both of these tasks, we follow community standards, with existing datasets.
%Note that many 3D reconstruction works (e.g.~\cite{OccupancyNetworks, DISN}) only reconstruct a surface without color and evaluate using 3D metrics. Our NeRF-based representation is capable of arbitrary view synthesis but does not explicitly construct a surface.
% To pursue a fair comparison across representations, we follow SRN~\cite{SRN} and use image quality metrics for comparing with recent works~\cite{SRN, ENR, DVR} which are capable of inferring object texture.

\begin{table}
\centering
    % https://docs.google.com/spreadsheets/d/1Zs5fO8UNJMRN74g_9JYAizGgxFKP2hg96JD3QkR5r6M/edit#gid=0
% \begin{tabular}{@{}lllll@{}}
% \toprule
%       & \multicolumn{2}{c}{1-view}                            & \multicolumn{2}{c}{2-view}                            \\
%       & \multicolumn{1}{c}{Chairs} & \multicolumn{1}{c}{Cars} & \multicolumn{1}{c}{Chairs} & \multicolumn{1}{c}{Cars} \\ \midrule
% SRN   & 22.89/0.89                 & 22.25/0.89               & 24.48/0.92                 & 24.84/0.92               \\
% ENR*  & 22.83/-                    & 22.26/-                  & -                          & -                        \\
% Ours* & \textbf{23.72/0.91}        & \textbf{23.17/0.90}      & \textbf{26.20/0.94}        & \textbf{25.66/0.94}      \\ \bottomrule
% \end{tabular}
\setlength{\tabcolsep}{5.0pt}
\begin{tabular}{@{}lllllll@{}}
\toprule
                    &                            & \multicolumn{2}{c}{1-view}                                                &                      & \multicolumn{2}{c}{2-view}                                                \\ \cmidrule(lr){3-4} \cmidrule(l){6-7} 
                    &  & \multicolumn{1}{c}{PSNR} & \multicolumn{1}{c}{SSIM} & \multicolumn{1}{c}{} & \multicolumn{1}{c}{PSNR} & \multicolumn{1}{c}{SSIM} \\ \midrule
\multirow{5}{*}{
    Chairs
}  

                    & GRF~\cite{GRF}                       & 21.25                               & 0.86                                   &                      & 22.65                                   & 0.88                                   \\
                    & TCO~\cite{Tatarchenko2015}~*                        & 21.27                               & 0.88                                &                      & 21.33                               & 0.88                                \\
                    & dGQN~\cite{GQN}                       & 21.59                               & 0.87                                &                      & 22.36                               & 0.89                                \\
                    & ENR~\cite{ENR}~*                       & 22.83                               & -                                   &                      & -                                   & -                                   \\
                    & SRN~\cite{SRN}                        & 22.89                               & 0.89                                &                      & 24.48                               & 0.92                                \\
                    & Ours~*                      & \textbf{23.72}                      & \textbf{0.91}                       &                      & \textbf{26.20}                      & \textbf{0.94}                       \\ \midrule
\multirow{3}{*}{
    Cars
}
                    & SRN~\cite{SRN}                        & 22.25                               & 0.89                                &                      & 24.84                               & 0.92                               \\
                    & ENR~\cite{ENR}~*                       & 22.26                               & -                                   &                      & -                                   & -                                   \\
                    & Ours~*                      & \textbf{23.17}                      & \textbf{0.90}                       &                      & \textbf{25.66}                      & \textbf{0.94}                       \\ \bottomrule
\end{tabular}
    \caption{
        \textbf{Category-specific 1- and 2-view reconstruction}.
        Methods marked * do not require canonical poses at test time.
        In all cases, a single model is trained for each category and used for both 1- and 2-view evaluation.
        Note ENR is a 1-view only model.
    }
    \label{tab:single_cat}
\end{table}

\begin{table}
    \centering
    \setlength{\tabcolsep}{1.0pt}
\begin{tabular}{@{}lcccccc@{}}
% \begin{tabular}{lllllll}
\toprule
                     & \multicolumn{3}{c}{1-view} & \multicolumn{3}{c}{2-view} \\ 
                    \cmidrule(lr){2-4} \cmidrule(l){5-7} 
                   & \multicolumn{1}{c}{$\uparrow$ PSNR} 
                   & \multicolumn{1}{c}{$\uparrow$ SSIM} 
                   & \multicolumn{1}{c}{$\downarrow$ LPIPS} 
                   & \multicolumn{1}{c}{$\uparrow$ PSNR} 
                   & \multicolumn{1}{c}{$\uparrow$ SSIM} 
                   & \multicolumn{1}{c}{$\downarrow$ LPIPS} \\
            \midrule
$-$~Local
& 20.39 & 0.848 & 0.196
& 21.17 & 0.865 & 0.175 \\
% freeze enc 
% & 18.77 & 0.793 & 0.285
% & 21.27 & 0.866 & 0.181 \\
$-$~Dirs
& 21.93 & 0.885 & 0.139
& 23.50 & 0.909 & 0.121 \\
Full
& \textbf{23.43} & \textbf{0.911} & \textbf{0.104}  
& \textbf{25.95} & \textbf{0.939} & \textbf{0.071} \\
\bottomrule
\end{tabular}
    \caption{\textbf{Ablation studies for ShapeNet chair reconstruction.} We show the benefit of using local features over a global code to condition the NeRF network ($-$Local vs Full), and of providing view directions to the network ($-$Dirs vs Full).
    }
    \vspace{-1em}
    \label{tab:chairs_ablations}
\end{table}

\begin{figure*}[ht]
\centering
    \includegraphics[width=\textwidth]{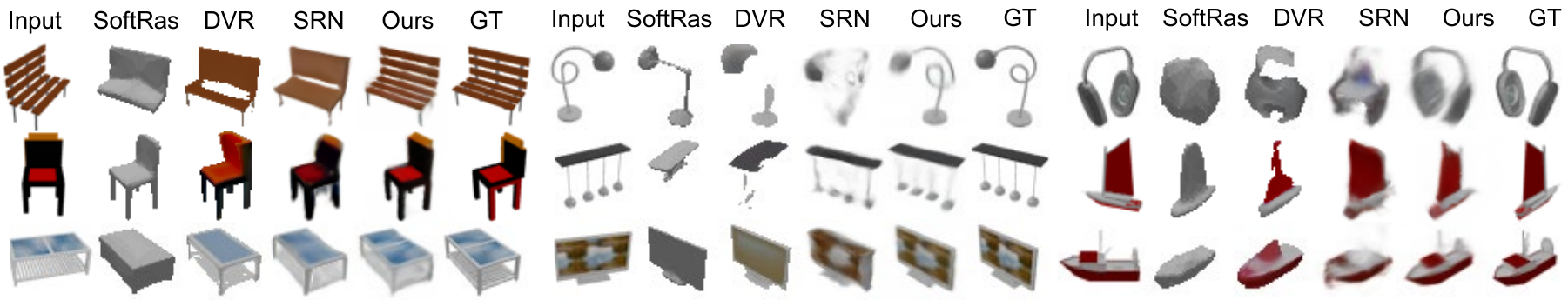}
\vspace{-0.6cm}
  \caption{\textbf{Category-agnostic single-view reconstruction.} 
  Going beyond the SRN benchmark, we train a single model to the 13 largest ShapeNet categories;
  we find that our approach produces superior visual results compared to a series of strong baselines.
  In particular, the model recovers fine detail and thin structure more effectively, even for outlier shapes.
  Quite visibly, images on monitors and tabletop textures are accurately reproduced; baselines
  representing the scene as a single latent vector cannot preserve
  such details of the input image.
  SRN's test-time latent inversion becomes less reliable as well in this setting.
  The corresponding quantitative evaluations are available in Table~\ref{tab:multi_cat}.
  Due to space constraints, we show objects with interesting  properties here.
  Please see the supplemental for sampled results.}
  \vspace{-.5em}
    \label{fig:category_agnostic}
\end{figure*}

\begin{table*}[t]
    \centering
\resizebox{\textwidth}{!}{
\begin{tabular}{@{}llllllllllllllll@{}}
\toprule
                   &      & plane          & bench          & cbnt.          & car            & chair          & disp.          & lamp           & spkr.          & rifle          & sofa           & table          & phone          & boat           & mean           \\ \midrule
\multirow{3}{*}{$\uparrow$ PSNR}    & DVR  & 25.29          & 22.64          & 24.47          & 23.95          & 19.91          & 20.86          & 23.27          & 20.78          & 23.44          & 23.35          & 21.53          & 24.18          & 25.09          & 22.70          \\
                   & SRN  & 26.62          & 22.20          & 23.42          & 24.40          & 21.85          & 19.07          & 22.17          & 21.04          & 24.95          & 23.65          & 22.45          & 20.87          & 25.86          & 23.28          \\
                   & Ours & \textbf{29.76} & \textbf{26.35} & \textbf{27.72} & \textbf{27.58} & \textbf{23.84} & \textbf{24.22} & \textbf{28.58} & \textbf{24.44} & \textbf{30.60} & \textbf{26.94} & \textbf{25.59} & \textbf{27.13} & \textbf{29.18} & \textbf{26.80} \\ \midrule
 \multirow{3}{*}{$\uparrow$ SSIM}  & DVR  & 0.905          & 0.866          & 0.877          & 0.909          & 0.787          & 0.814          & 0.849          & 0.798          & 0.916          & 0.868          & 0.840          & 0.892          & 0.902          & 0.860          \\
                   & SRN  & 0.901          & 0.837          & 0.831          & 0.897          & 0.814          & 0.744          & 0.801          & 0.779          & 0.913          & 0.851          & 0.828          & 0.811          & 0.898          & 0.849          \\
                   & Ours & \textbf{0.947} & \textbf{0.911} & \textbf{0.910} & \textbf{0.942} & \textbf{0.858} & \textbf{0.867} & \textbf{0.913} & \textbf{0.855} & \textbf{0.968} & \textbf{0.908} & \textbf{0.898} & \textbf{0.922} & \textbf{0.939} & \textbf{0.910} \\ \midrule
 \multirow{3}{*}{$\downarrow$ LPIPS} & DVR  & 0.095          & 0.129          & 0.125          & 0.098          & 0.173          & 0.150          & 0.172          & 0.170          & 0.094          & 0.119          & 0.139          & 0.110          & 0.116          & 0.130          \\
                   & SRN  & 0.111          & 0.150          & 0.147          & 0.115          & 0.152          & 0.197          & 0.210          & 0.178          & 0.111          & 0.129          & 0.135          & 0.165          & 0.134          & 0.139          \\
                   & Ours & \textbf{0.084} & \textbf{0.116} & \textbf{0.105} & \textbf{0.095} & \textbf{0.146} & \textbf{0.129} & \textbf{0.114} & \textbf{0.141} & \textbf{0.066} & \textbf{0.116} & \textbf{0.098} & \textbf{0.097} & \textbf{0.111} & \textbf{0.108} \\ \bottomrule
\end{tabular}
}
\vspace{-0.01cm}
    
    \caption{
        \textbf{Category-agnostic single-view reconstruction.} 
        Quantitative results for category-agnostic view-synthesis are presented, with a detailed breakdown by category. 
        Our method outperforms the state-of-the-art by significant margins in all categories.
    }
    \label{tab:multi_cat}
    \vspace{-1em}
\end{table*}

\vspace{-0.5em}
\subsubsection{Category-specific View Synthesis Benchmark}
\label{sec:single_cat}
We perform one-shot and two-shot view synthesis on the ``chair" and ``car" classes of ShapeNet, using the protocol and dataset introduced in~\cite{SRN}.
The dataset contains 6591 chairs and 3514 cars with a predefined split across object instances.
All images have resolution $128 \times 128$.

A single model is trained for each object class with 50 random views per object instance, randomly sampling either one or two of the training views to encode.
For testing, 
We use 251 novel views on an Archimedean spiral for each object in the test set of object instances, fixing 1-2 informative views as input.
We report our performance in comparison with state-of-the-art baselines in Table~\ref{tab:single_cat}, and show selected qualitative results in Fig.~\ref{fig:single_view_cars_chairs}.
We also include the quantitative results of baselines TCO~\cite{Tatarchenko2015}
and dGQN~\cite{GQN} reported in~\cite{SRN} where applicable, and
the values available in the recent works ENR~\cite{ENR} and GRF~\cite{GRF} in this setting.
% \interfootnotelinepenalty=10000
% ~\footnote{We (and ENR \cite{ENR}) use a modified dataset provided by the SRN authors, and the numbers reported in~\cite{SRN, GRF} no longer apply for cars. For \cite{ENR, GRF} no source code is yet released, so we report only numbers from the papers.}

PixelNeRF achieves noticeably superior results despite solving a problem \textit{significantly harder} than SRN because we:
 \begin{inparaenum}[1)]
     \item use feed-forward prediction, without test-time optimization,
     \item do not use ground-truth absolute camera poses at test-time,
     \item use view instead of canonical space.
 \end{inparaenum}

\vspace{0.5em}
\noindent\textbf{Ablations.}
 In Table~\ref{tab:chairs_ablations},
we show the benefit of using local features and view directions in our model for this category-specific setting.
Conditioning the NeRF network on pixel-aligned local features instead of a global code ($-$Local vs Full) improves performance significantly, for both single and two-view settings.
Providing view directions ($-$Dirs vs Full) also provides a significant boost.
For these ablations, we follow an abbreviated evaluation protocol on ShapeNet chairs, using 25 novel views on the Archimedean spiral.

\vspace{-0.6em}
\subsubsection{Category-agnostic Object Prior}
\label{sec:multi_cat}
While we found appreciable improvements over baselines in the simplest category-specific benchmark,
our method is by no means constrained to it. 
We show in Table~\ref{tab:multi_cat} and 
Fig.~\ref{fig:category_agnostic} that
our approach offers a much greater advantage in the
\textit{category-agnostic} setting of \cite{SoftRas, DVR}, where we train a single model to the $13$ largest categories of ShapeNet.
%We used pretrained SoftRas and DVR models and 
%trained a new SRN model for 14 days on an RTX Titan for comparison.
Please see the supplemental for randomly sampled results.

We follow community standards for 2D-supervised methods on multiple ShapeNet categories \cite{DVR, NMR, SoftRas} and use the renderings and splits from Kato et al.~\cite{NMR}, which provide 
24 fixed elevation views of $64 \times 64$ resolution for each object instance.
% which consists of $24$ images of size $64 \times 64$ with fixed elevation for each object.
During both training and evaluation, a random view is selected as the input view for each object and shared across all baselines.
% where this input view selection is also shared across all baselines.
The remaining $23$ views are used as target views for computing metrics (see
$\S$~\ref{sec:metrics}).

\subsection{Pushing the Boundaries of ShapeNet}
\label{sec:beyond_shapenet}
%\ak{Briefly explain what they are in keywords: generalization to novel categories, multiple objects, and evaluation on real images.}
Taking a step towards reconstruction in less controlled capture scenarios, we perform experiments on ShapeNet data in three more challenging setups:
\begin{inparaenum}[1)]
 \item unseen object categories,
 \item multiple-object scenes, and
 \item simulation-to-real transfer on car images.
\end{inparaenum}
In these settings, successful reconstruction requires geometric priors; recognition or retrieval alone is not sufficient.

\begin{figure}
\centering
\includegraphics[width=\linewidth]{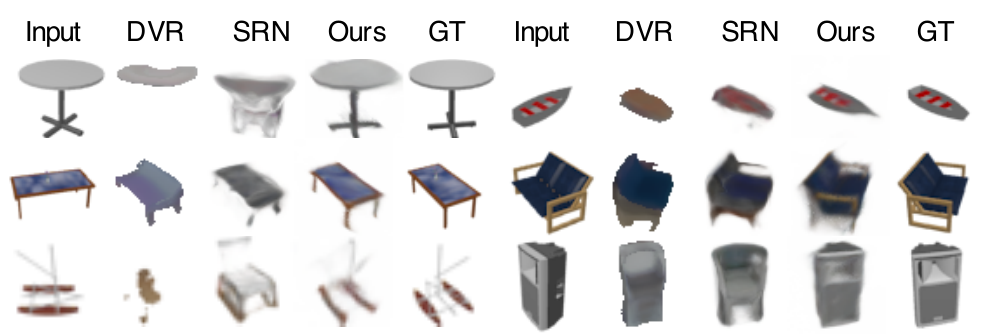}
    \vspace{-1.3em}
   \caption{
        \textbf{Generalization to unseen categories.}
        We evaluate a model trained on planes, cars, and chairs on 10 unseen ShapeNet categories.
        We find that the model is able to synthesize reasonable views even in this difficult case.}
    \vspace{-0.1em}
        \label{fig:novel_cat}
\end{figure}

\begin{figure}
\begin{center}
\includegraphics[width=\linewidth]{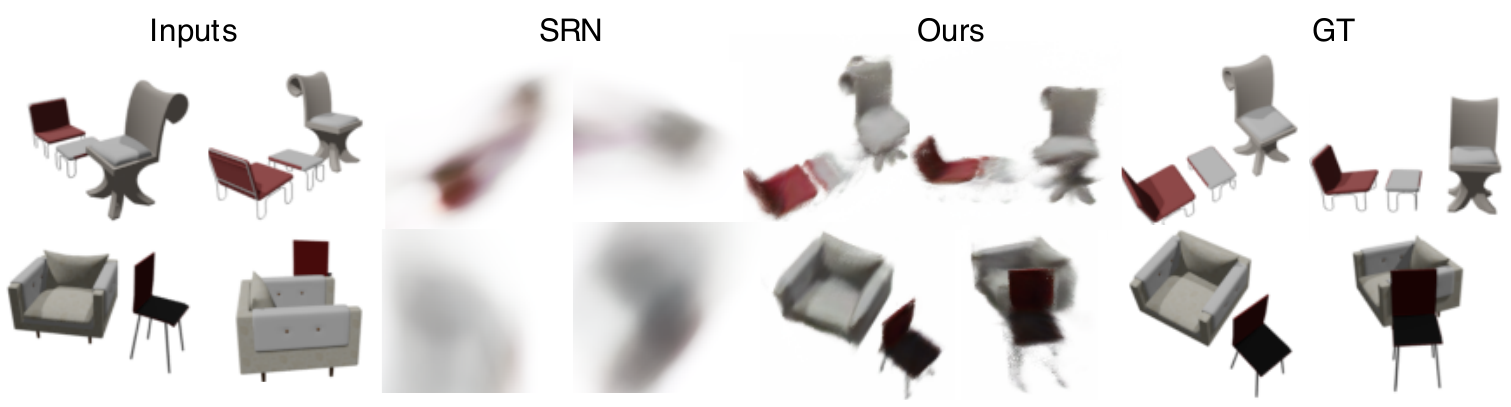}
\vspace{-8mm}
\end{center}
   \caption{\textbf{360\degree view prediction with multiple objects.}
   We show qualitative results of our method compared with SRN on scenes composed of multiple ShapeNet chairs. We are easily able to handle this setting, because our prediction is done in view space; in contrast, SRN predicts in canonical space, and struggles with scenes that cannot be aligned in such a way.
   }
   \vspace{-.3em}
   \label{fig:2obj}
\end{figure}

\begin{table}
\centering
    \setlength{\tabcolsep}{1.5pt}
\begin{tabular}{@{}lcccccc@{}}
% \begin{tabular}{lllllll}
\toprule
                     & \multicolumn{3}{c}{Unseen category} & \multicolumn{3}{c}{Multiple chairs} \\ 
                    \cmidrule(lr){2-4} \cmidrule(l){5-7} 
                   & \multicolumn{1}{c}{$\uparrow$ PSNR} 
                   & \multicolumn{1}{c}{$\uparrow$ SSIM} 
                   & \multicolumn{1}{c}{$\downarrow$ LPIPS} 
                   & \multicolumn{1}{c}{$\uparrow$ PSNR} 
                   & \multicolumn{1}{c}{$\uparrow$ SSIM} 
                   & \multicolumn{1}{c}{$\downarrow$ LPIPS} \\
            \midrule
DVR  
& 17.72 & 0.716 & 0.240
& - & - & - \\
SRN  
& 18.71 & 0.684 & 0.280
& 14.67 & 0.664 & 0.431 \\
Ours 
& \textbf{22.71} & \textbf{0.825} & \textbf{0.182} 
& \textbf{23.40} & \textbf{0.832} & \textbf{0.207} \\ 
\bottomrule
\end{tabular}
    \caption{
        \textbf{Image quality metrics for challenging ShapeNet tasks.}
        (Left) Average metrics on 10 unseen categories for models trained on only planes, cars, and chairs. 
        See the supplemental for a breakdown by category. 
        (Right) Average metrics for two-view reconstruction for scenes with multiple ShapeNet chairs.
    }
    \vspace{-1em}
    \label{tab:beyond_shapenet}
\end{table}

\begin{figure}
    \centering
    \includegraphics[width=1\linewidth]{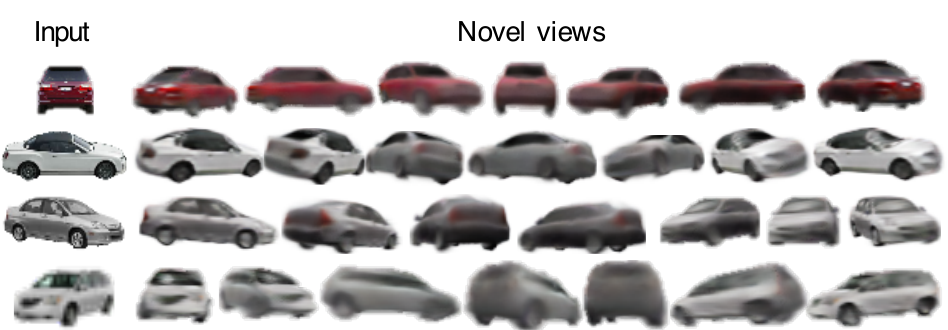}
   \caption{\textbf{Results on real car photos.} 
   We apply the 
   car model from $\S$~\ref{sec:single_cat} directly to images from the Stanford cars dataset~\cite{CarsDataset}. The background has been masked out using PointRend~\cite{PointRend}.
   The views are rotations about the view-space vertical axis.}
   \label{fig:real_car}
   \vspace{-1em}
\end{figure}

\begin{figure*}[t]
\centering
    \includegraphics[width=\linewidth]{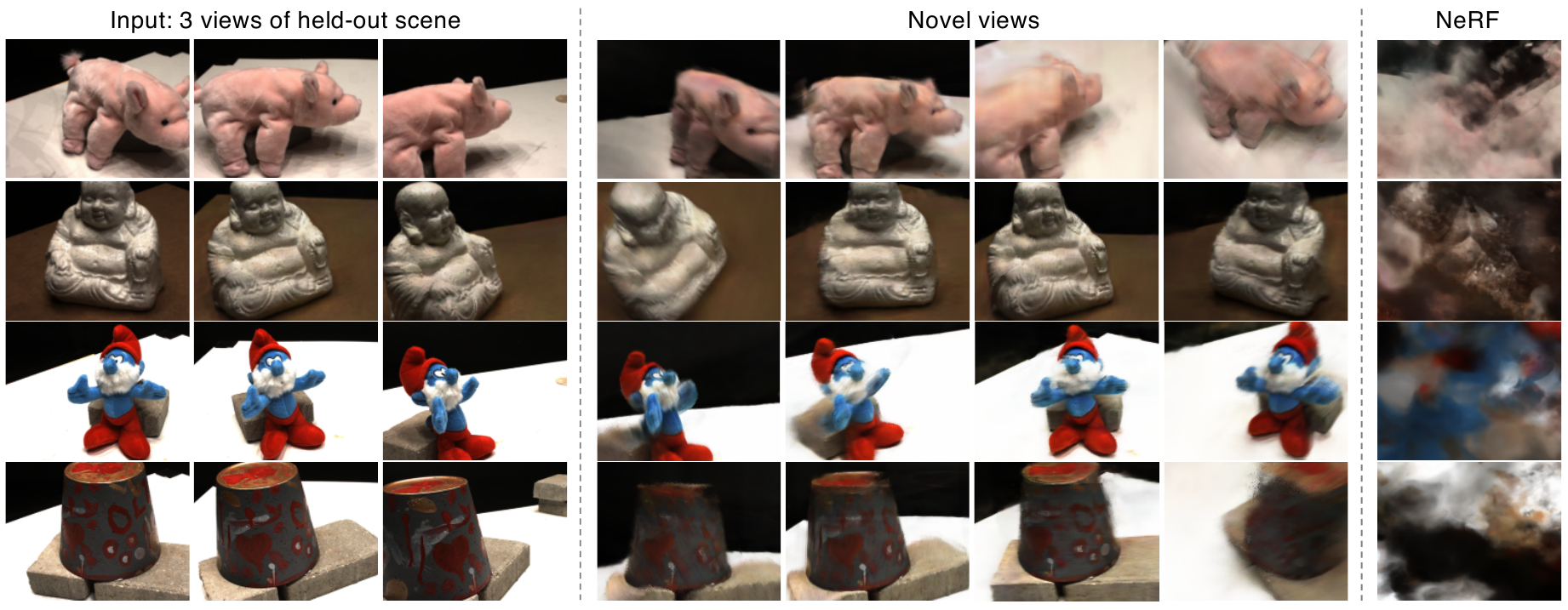}
    \vspace{-1.5em}
   \caption{\textbf{Wide baseline novel-view synthesis on a real image dataset.} 
   We train our model to distinct scenes in the DTU MVS dataset~\cite{DTU}.
   Perhaps surprisingly,  even in this case, our model is able to infer novel views with reasonable quality for held-out scenes 
   without further test-time optimization, all from only three views.
   Note the train/test sets share no overlapping scenes.}
  \vspace{-1em}
   \label{fig:dtu}
\end{figure*}

\vspace{0.5em}
\noindent\textbf{Generalization to novel categories.}
\label{sec:cross_cat_gen}
We first aim to reconstruct ShapeNet categories which were not seen in training.
Unlike the more standard category-agnostic task described in the previous section, such generalization is impossible with semantic information alone.
The results in Table~\ref{tab:beyond_shapenet} and Fig.~\ref{fig:novel_cat}
suggest our method learns intrinsic geometric and appearance priors which 
are fairly effective
even for objects quite distinct from those seen during training.

We loosely follow the protocol used for zero-shot cross-category reconstruction from~\cite{GenRe, yan2017perspective}.
Note that our baselines ~\cite{SRN, DVR} do not evaluate in this setting, and we adapt them for the sake of comparison.
We train on the airplane, car, and chair categories and test on 10 categories unseen during training, continuing to use the Kato et al. renderings described in $\S$~\ref{sec:multi_cat}.

\vspace{0.5em}
\noindent\textbf{Multiple-object scenes.}
We further perform few-shot $360\degree$ reconstruction for scenes with multiple randomly placed and oriented ShapeNet chairs.
In this setting, the network cannot rely solely on semantic cues for correct object placement and completion.
The priors learned by the network must be applicable in an arbitrary coordinate system.
We show in Fig.~\ref{fig:2obj} and Table~\ref{tab:beyond_shapenet} that our formulation allows us to perform well on these simple scenes without additional design modifications.
In contrast, SRN models scenes in a canonical space and struggles on held-out scenes.

% This setting is more difficult because objects can be arbitrarily oriented with respect to each other.
% This makes it difficult for a network to rely solely on semantic cues, and requires that the geometric priors learned by the network be applicable in an arbitrary coordinate system.
We generate training images composed with 20 views randomly sampled on the hemisphere and
render test images composed of a held out test set of chair instances, with 50 views sampled on an Archimedean spiral.
During training, we randomly encode two input views; at test-time, we fix two informative views across the compared methods.
In the supplemental, we provide example images from our dataset as well as additional quantitative results and qualitative comparisons with varying numbers of input views.

\vspace{0.4em}
\noindent\textbf{Sim2Real on Cars.}
We also explore the performance of pixelNeRF on real images from the Stanford cars dataset~\cite{CarsDataset}.
We directly apply car model from $\S$~\ref{sec:single_cat} without any fine-tuning.
As seen in Fig.~\ref{fig:real_car}, the network trained on synthetic data effectively infers shape and texture of the real cars, suggesting our model can transfer beyond the synthetic domain.

Synthesizing the $360\degree$ background from a single view is nontrivial and out of the scope for this work.
For this demonstration, the off-the-shelf PointRend~\cite{PointRend} segmentation model is used to remove
the background.

%We hypothesize that the degradation in visual sharpness for more far-away views is
%partly due to that ShapeNet cars almost exclusively have Lambertian materials.

\begin{figure}
    \centering
    \includegraphics[width=0.9\linewidth]{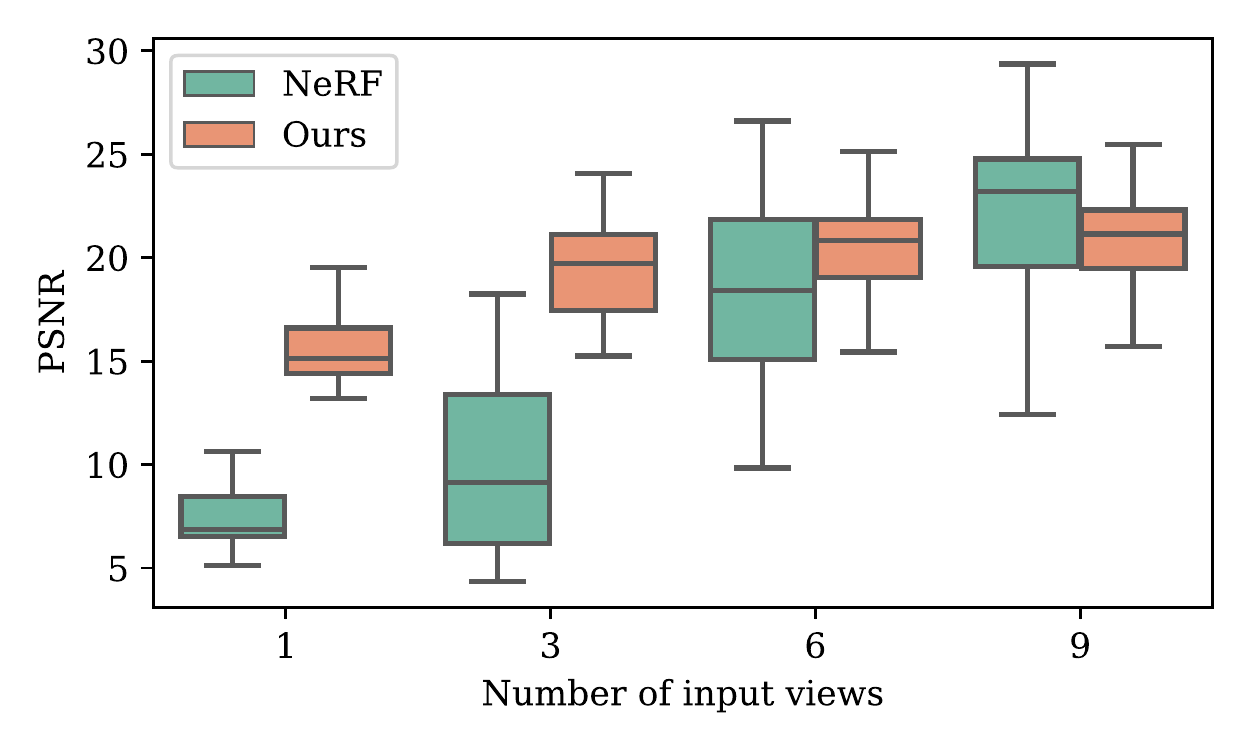}
    \vspace{-0.5em}
    \caption{\textbf{PSNR of few-shot feed-forward DTU reconstruction.}
    We show the quantiles of PSNR on DTU for our method and NeRF, given 1, 3, 6, or 9 input views.
    Separate NeRFs are trained per scene and number of input views, while our method requires only a single model trained with 3 encoded views.}
    \label{fig:dtu_views}
    \vspace{-1.5em}
\end{figure}
\subsection{Scene Prior on Real Images}
\label{sec:dtu}
Finally, we demonstrate that our method is applicable for few-shot \textit{wide baseline} novel-view synthesis on real scenes in the DTU MVS dataset~\cite{DTU}. 
Learning a prior for view synthesis on this dataset poses significant challenges:
not only does it consist of more complex scenes, without clear semantic similarities across scenes, it also contains inconsistent backgrounds and lighting between scenes.
Moreover, under 100 scenes are available for training.
% We note that previous work~\cite{DVR, IDR} use manually annotated object masks on a small subset of this dataset.
% Our method does not require mask supervision and can use a larger number of scenes to better learn a scene prior.
We found that the standard data split introduced in MVSNet~\cite{MVSNet} contains overlap between scenes of the training and test sets.
Therefore, for our purposes, we use a different split of 88 training scenes and 15 test scenes, in which there are no shared or highly similar scenes between the two sets.
% We omit the remaining 21 scenes because of similarity to existing test scenes, or low capture quality. 
Images are down-sampled to a resolution of $400 \times 300$.

We train one model across all training scenes by encoding $3$ random views of a scene.
During test time, we choose a set of fixed informative input views shared across all instances.
We show in Fig.~\ref{fig:dtu} that our method can perform view synthesis on the held-out test scenes. 
We further quantitatively compare the performance of our feed-forward model with NeRF optimized to the same set of input views
in Fig.~\ref{fig:dtu_views}.
Note that training each of 60 NeRFs took 14 hours; in contrast,
pixelNeRF is applied to new scenes immediately without any test-time optimization.
\section{Discussion}

We have presented pixelNeRF, a framework to learn a scene prior for reconstructing NeRFs from one or a few images.
Through extensive experiments, we have established that our approach can be successfully applied in a variety of settings.
We addressed some shortcomings of NeRF, but there are challenges yet to be explored:
\begin{inparaenum}[1)]
\item
Like NeRF, our rendering time is slow, and in fact,
our runtime increases linearly when given more input views.
Further, some methods (e.g.~\cite{DVR, SoftRas}) can recover a mesh from  the image enabling fast rendering and manipulation afterwards, while NeRF-based representations cannot be converted to meshes very reliably.
Improving NeRF's efficiency is an
important research question that
can enable real-time applications.
\item
As in the vanilla NeRF,
we manually tune ray sampling bounds $t_n, t_f$ and a scale for the positional encoding.
Making NeRF-related methods scale-invariant is a crucial challenge.
\item
While we have demonstrated our method on real data from the DTU dataset, we acknowledge that 
this dataset was captured under controlled settings and has matching camera poses across all scenes with limited viewpoints.
Ultimately, our approach is bottlenecked by the availability of large-scale wide baseline multi-view datasets, limiting the applicability to datasets such as ShapeNet and DTU.  
Learning a general prior for $360\degree$ scenes in-the-wild is an exciting direction for future work.
\end{inparaenum}

%We make the common assumption of an \textit{a priori} fixed focal length, which does not correctly handle perspective distortion. 
%Our method does not leverage symmetry as well as canonical-space methods to 
% complete the non-visible parts where the image features are not informative
\FloatBarrier

\section*{Acknowledgements}
 We thank Shubham Goel and Hang Gao for comments on the text. We also thank Emilien Dupont and Vincent Sitzmann for helpful discussions.

{\small
\bibliographystyle{ieee_fullname}
\bibliography{egbib}
}

% APPENDIX
% Removed temporarily bc it is messing with reference numbers

\renewcommand{\thesection}{\Alph{section}}

\setcounter{section}{0}
\section*{Appendix}

\section{Additional Results}
In this section, we provide additional qualitative and quantitative results
for several key experiments.
The reader is encouraged to 
refer to the video and website for 
a richer, animated presentation of qualitative results.

\subsection{Category-agnostic ShapeNet: Random Results}
\begin{figure*}[t]
    \begin{center}
    \includegraphics[width=1.0\linewidth]{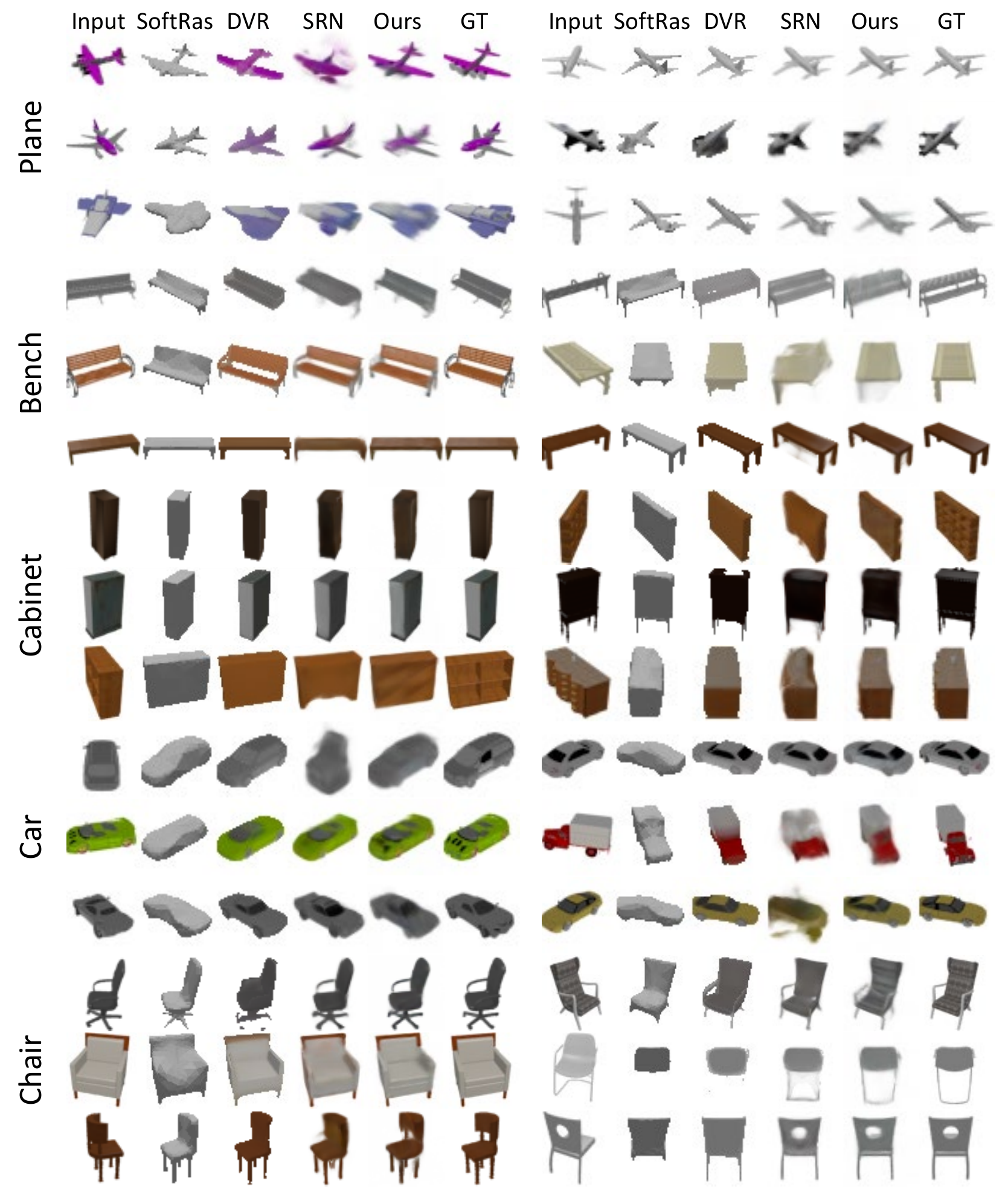}
    \end{center}
    \caption{\textbf{Randomly sampled results.} part 1}
    \label{fig:random_1}
\end{figure*}

\begin{figure*}[t]
    \begin{center}
    \includegraphics[width=1.0\linewidth]{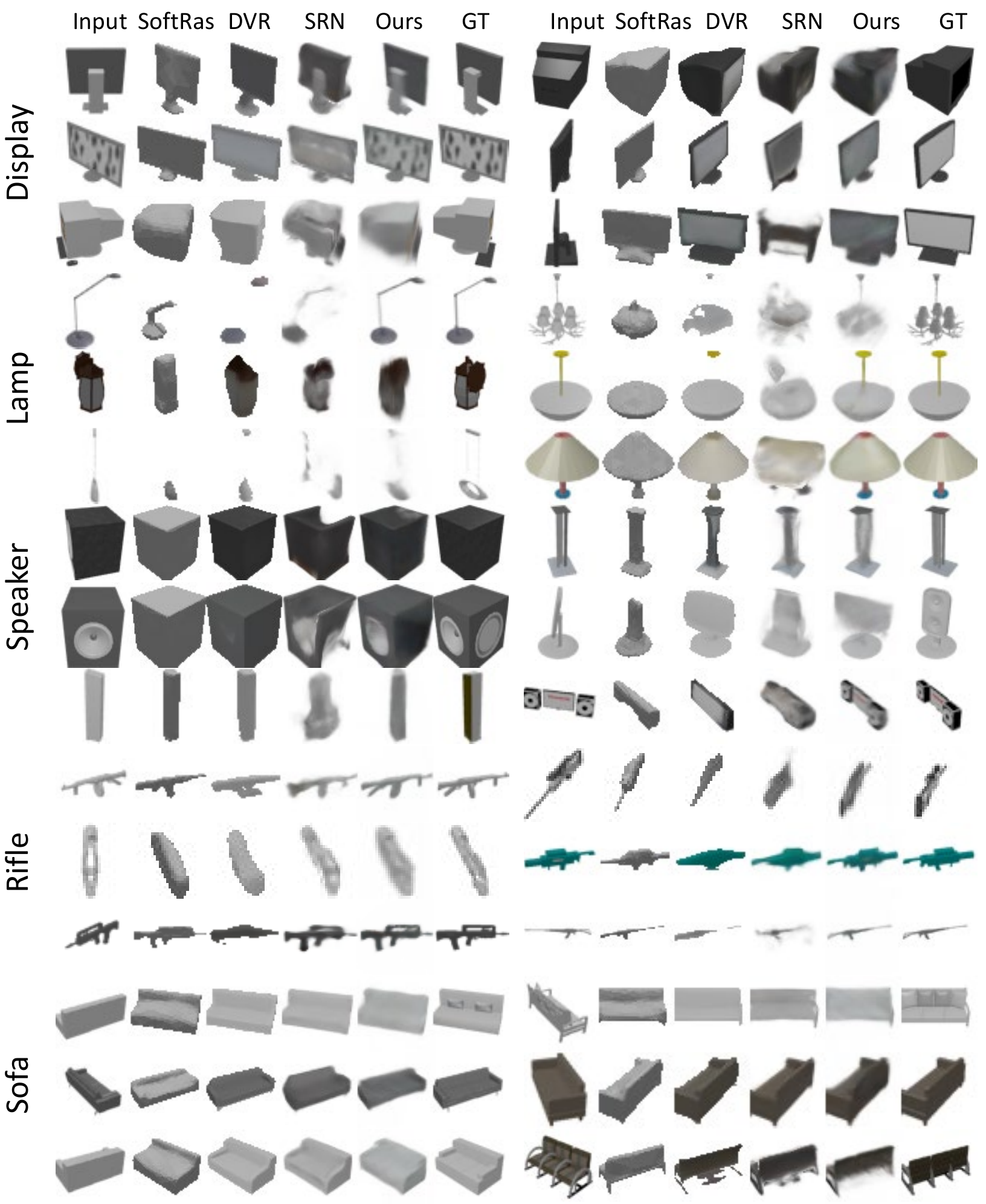}
    \end{center}
    \caption{\textbf{Randomly sampled results.} part 2}
    \label{fig:random_2}
    \vspace{-0.5em}
\end{figure*}

\begin{figure*}[t]
    \centering
    \begin{center}
    \includegraphics[width=1.0\linewidth]{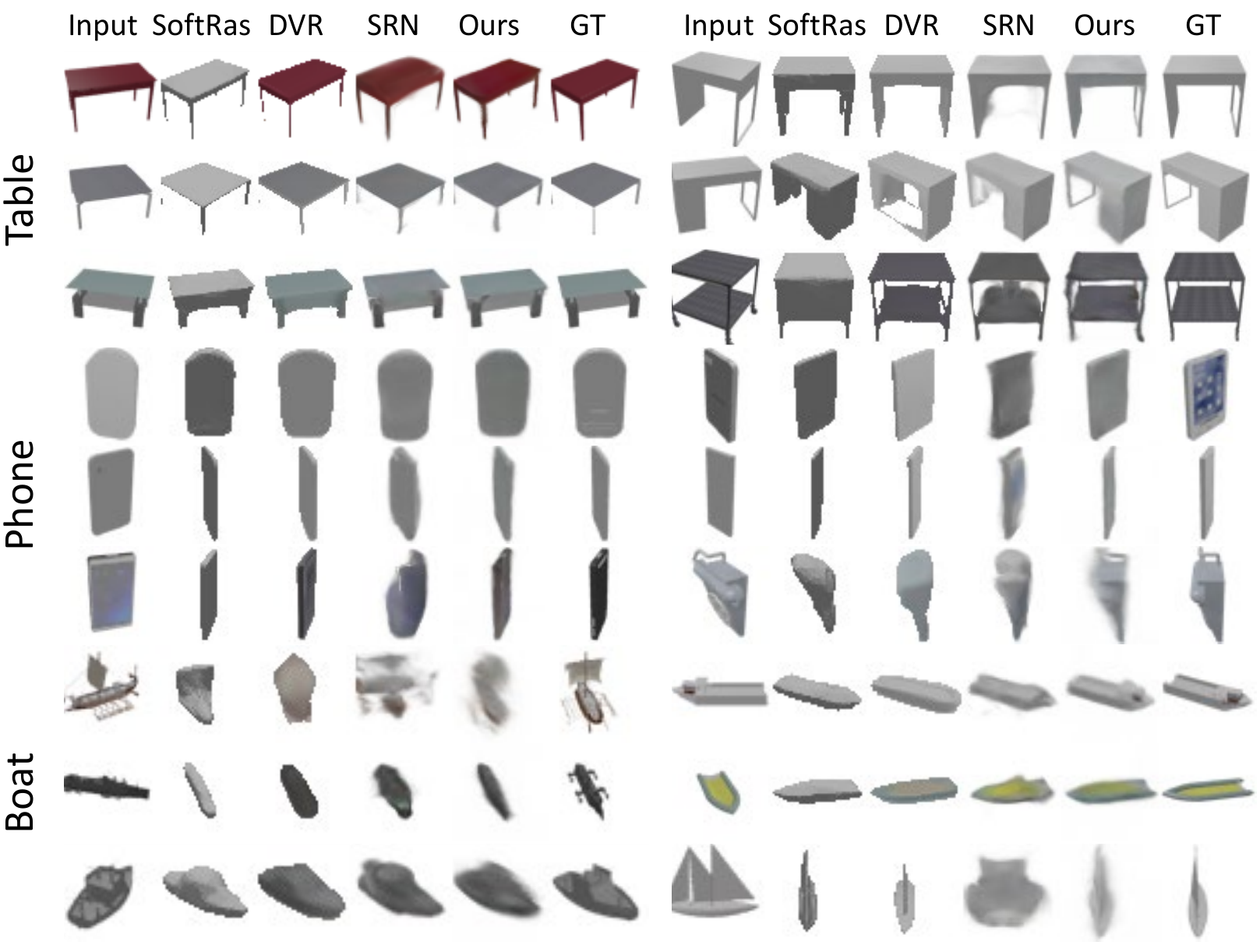}
    \end{center}
    \caption{\textbf{Randomly sampled results.} part 3}
    \label{fig:random_3}
\end{figure*}

We show \emph{randomly} sampled results 
for the category-agnostic setting 
($\S$~\ref{sec:multi_cat})
in Fig.~\ref{fig:random_1},
Fig.~\ref{fig:random_2}, and Fig.~\ref{fig:random_3}.
Specifically, we sample $6$ uniformly random objects
for each of the 13 largest ShapeNet categories and show comparisons to the baselines~\cite{SoftRas, DVR, SRN} as in the main paper.
Two random views are selected from the $24$ available views 
to be source and target views respectively.

\subsection{Generalization to novel categories}

In Table~\ref{tab:gen_cat_breakdown} we show a detailed breakdown of metrics by category on
unseen categories, as promised in the main paper.

\begin{table*}[t]
    \vspace{2em}
    \begin{center}
    \begin{tabular}{@{}lllllllllllll@{}}
\toprule
                                                              &      & bench          & cbnt.          & disp.          & lamp           & spkr.          & rifle          & sofa           & table          & phone          & boat           & mean           \\ \midrule
\multirow{3}{*}{$\uparrow$ PSNR}    & DVR  & 18.37          & 17.19          & 14.33          & 18.48          & 16.09          & 20.28          & 18.62          & 16.20          & 16.84          & 22.43          & 17.72          \\
                                                              & SRN  & 18.71          & 17.04          & 15.06          & 19.26          & 17.06          & 23.12          & 18.76          & 17.35          & 15.66          & 24.97          & 18.71          \\
                                                              & Ours & \textbf{23.79} & \textbf{22.85} & \textbf{18.09} & \textbf{22.76} & \textbf{21.22} & \textbf{23.68} & \textbf{24.62} & \textbf{21.65} & \textbf{21.05} & \textbf{26.55} & \textbf{22.71} \\\midrule
\multirow{3}{*}{$\uparrow$ SSIM}    & DVR  & 0.754          & 0.686          & 0.601          & 0.749          & 0.657          & 0.858          & 0.755          & 0.644          & 0.731          & 0.857          & 0.716          \\
                                                              & SRN  & 0.702          & 0.626          & 0.577          & 0.685          & 0.633          & 0.875          & 0.702          & 0.617          & 0.635          & 0.875          & 0.684          \\
                                                              & Ours & \textbf{0.863} & \textbf{0.814} & \textbf{0.687} & \textbf{0.818} & \textbf{0.778} & \textbf{0.899} & \textbf{0.866} & \textbf{0.798} & \textbf{0.801} & \textbf{0.896} & \textbf{0.825} \\\midrule
\multirow{3}{*}{$\downarrow$ LPIPS} & DVR  & 0.219          & 0.257          & 0.306          & 0.259          & 0.266          & 0.158          & 0.196          & 0.280          & 0.245          & 0.152          & 0.240          \\
                                                              & SRN  & 0.282          & 0.314          & 0.333          & 0.321          & 0.289          & 0.175          & 0.248          & 0.315          & 0.324          & 0.163          & 0.280          \\
                                                              & Ours & \textbf{0.164} & \textbf{0.186} & \textbf{0.271} & \textbf{0.208} & \textbf{0.203} & \textbf{0.141} & \textbf{0.157} & \textbf{0.188} & \textbf{0.207} & \textbf{0.148} & \textbf{0.182} \\ \bottomrule
\end{tabular}
    \end{center}
    \caption{
        \textbf{Generalization to novel categories}.
        Expanding on Table~\ref{tab:beyond_shapenet} in the main paper,
        we show quantitative results with a 
        breakdown by category. 
    }
    \label{tab:gen_cat_breakdown}
    \vspace{2em}
\end{table*}

\subsection{Two-object Scenes}
\begin{figure*}
    \vspace{1.5em}
    \centering
    \begin{subfigure}[t]{0.8\textwidth}
        \centering
        \includegraphics[width=\linewidth]{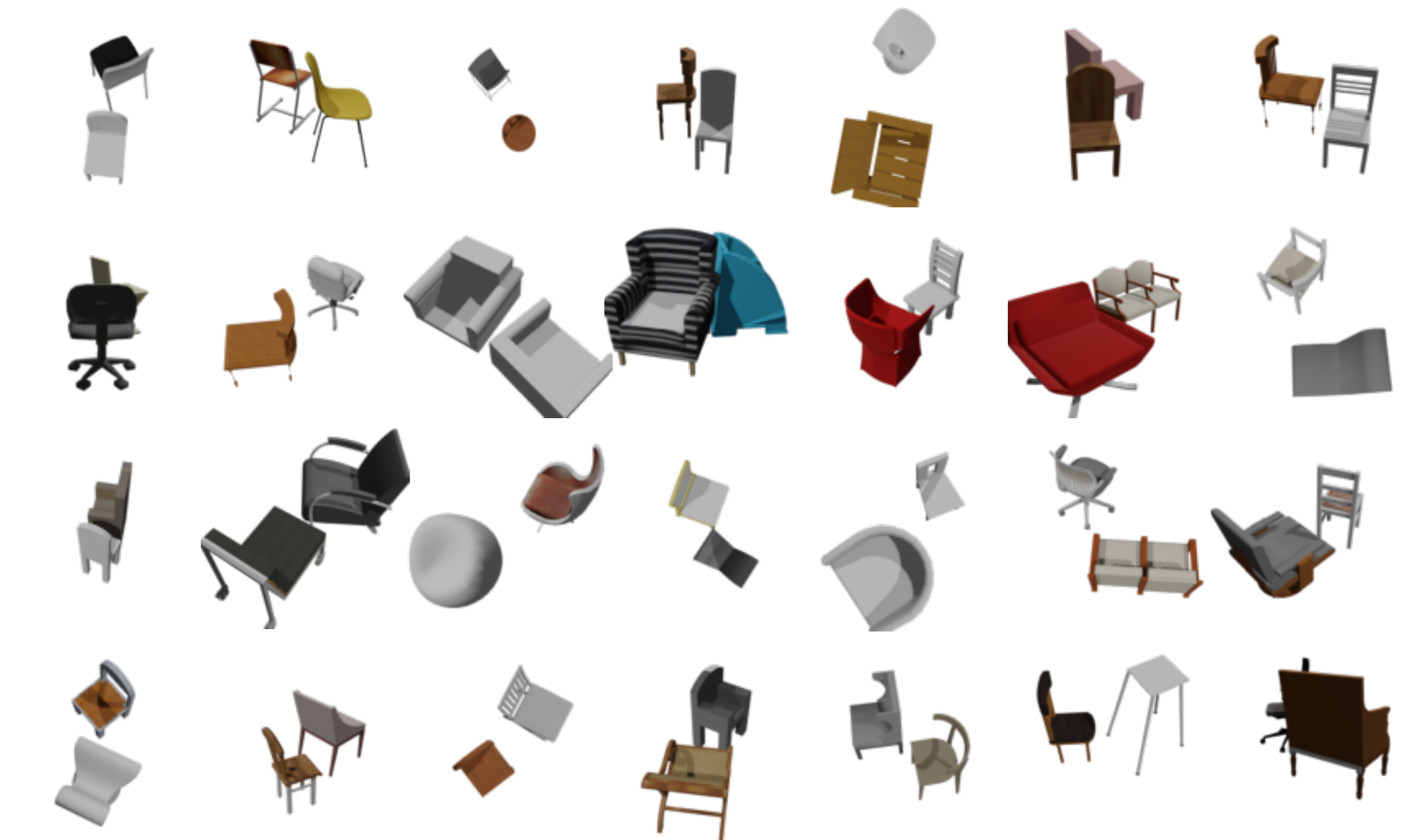}
        \caption{Train set}
    \end{subfigure}%
    \\
    \begin{subfigure}[t]{0.8\textwidth}
        \centering
        \includegraphics[width=\linewidth]{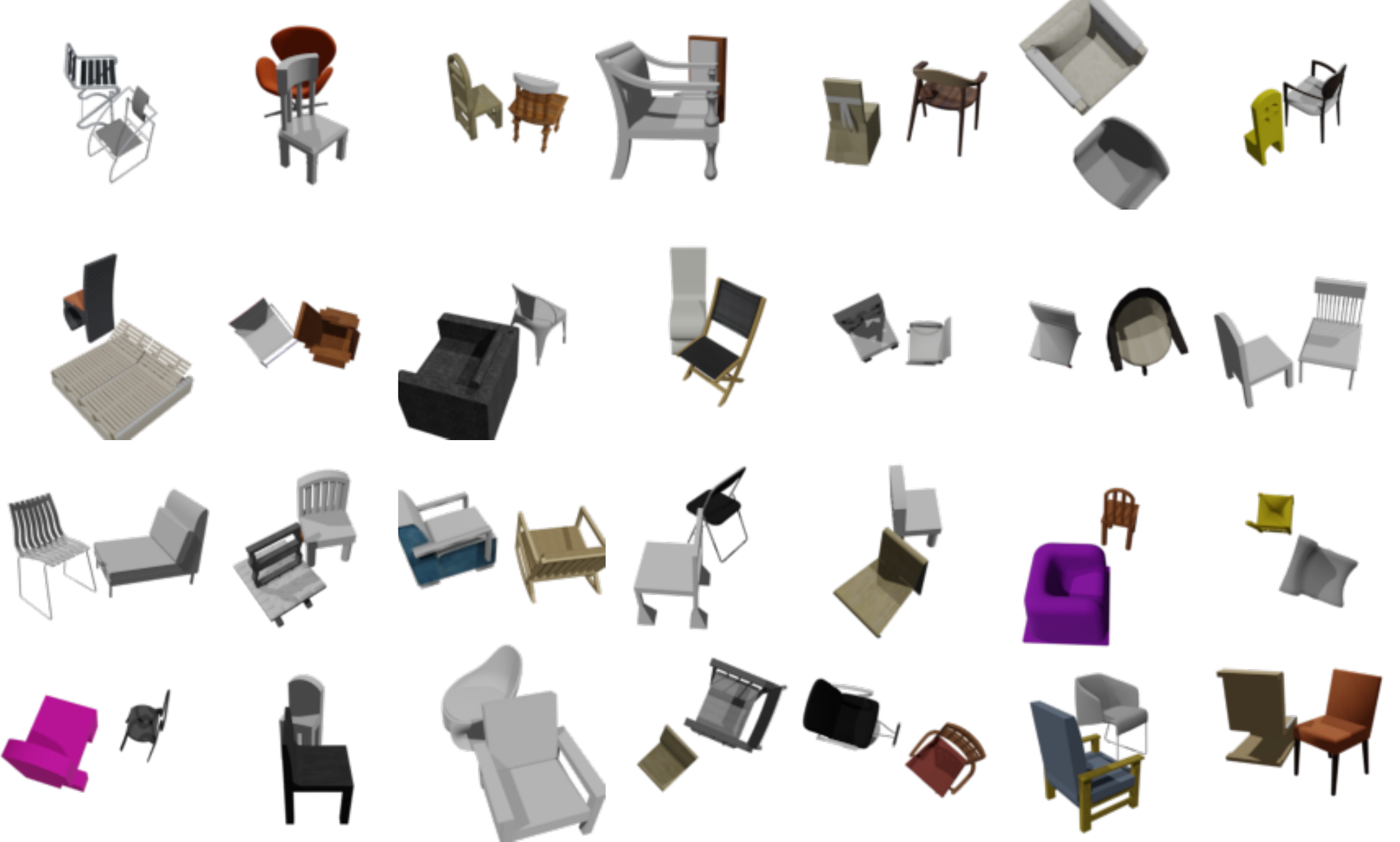}
        \caption{Test set}
    \end{subfigure}%
    \caption{Randomly sampled images from the synthetic two-object scene dataset}
    \label{fig:sample_2obj}
    \vspace{1em}
\end{figure*}

\begin{figure*}
    \centering
    \includegraphics[width=\textwidth]{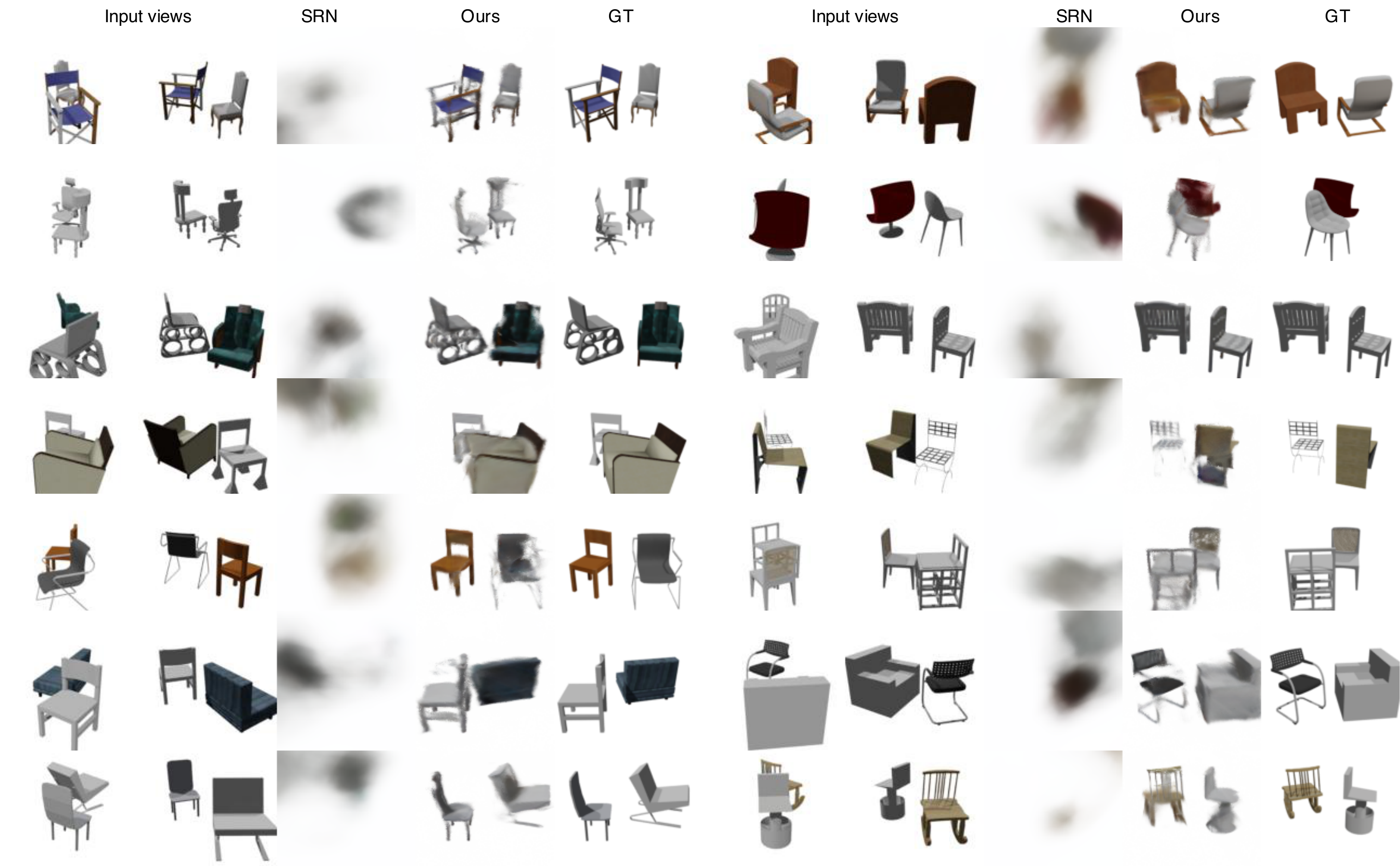}
    \caption{Randomly sampled results for two object scenes, when given two input views.}
    \label{fig:2obj_random}
\end{figure*}

We show samples from our rendered dataset in
Fig.~\ref{fig:sample_2obj}.  
An analysis of performance as more views become available is
in Table~\ref{tab:2obj_views}, for our method when compared with SRN.
We also show \textit{randomly sampled} results of scenes when given two input views in Figure~\ref{fig:2obj_random}.
We train our model using two random views, and give the model either one, two, or three fixed informative views during inference.

\begin{table*}
    \centering
    \setlength{\tabcolsep}{6pt}
\begin{tabular}{@{}lccccccccc@{}}
% \begin{tabular}{lllllll}
\toprule
                     & \multicolumn{3}{c}{1-view} & \multicolumn{3}{c}{2-view} &
                     \multicolumn{3}{c}{3-view}\\ 
                    \cmidrule(lr){2-4} \cmidrule(l){5-7} \cmidrule(l){8-10} 
                   & \multicolumn{1}{c}{$\uparrow$ PSNR} 
                   & \multicolumn{1}{c}{$\uparrow$ SSIM} 
                   & \multicolumn{1}{c}{$\downarrow$ LPIPS} 
                   & \multicolumn{1}{c}{$\uparrow$ PSNR} 
                   & \multicolumn{1}{c}{$\uparrow$ SSIM} 
                   & \multicolumn{1}{c}{$\downarrow$ LPIPS}
                    & \multicolumn{1}{c}{$\uparrow$ PSNR} 
                   & \multicolumn{1}{c}{$\uparrow$ SSIM} 
                   & \multicolumn{1}{c}{$\downarrow$ LPIPS}\\
            \midrule
SRN  
&13.76	& 0.658 & 0.422 
& 14.28 &	0.660	 & 0.432
& 14.67 &	0.664 &	0.431 \\
Ours 
& \textbf{20.15} &	\textbf{0.767} &	\textbf{0.274}
& \textbf{23.40} &	\textbf{0.832} &	\textbf{0.207}
& \textbf{23.68} &	\textbf{0.800} &	\textbf{0.206} \\ 
\bottomrule
\end{tabular}
    \caption{\textbf{Performance on synthetic two-object dataset with increasing number of views at test time.} Image quality metrics for SRN and our method, when increasing the number of views given at test time.}
    \label{tab:2obj_views}
\end{table*}

\subsection{DTU}

\begin{figure*}[t]
    \centering
    \includegraphics[width=1.0\linewidth]{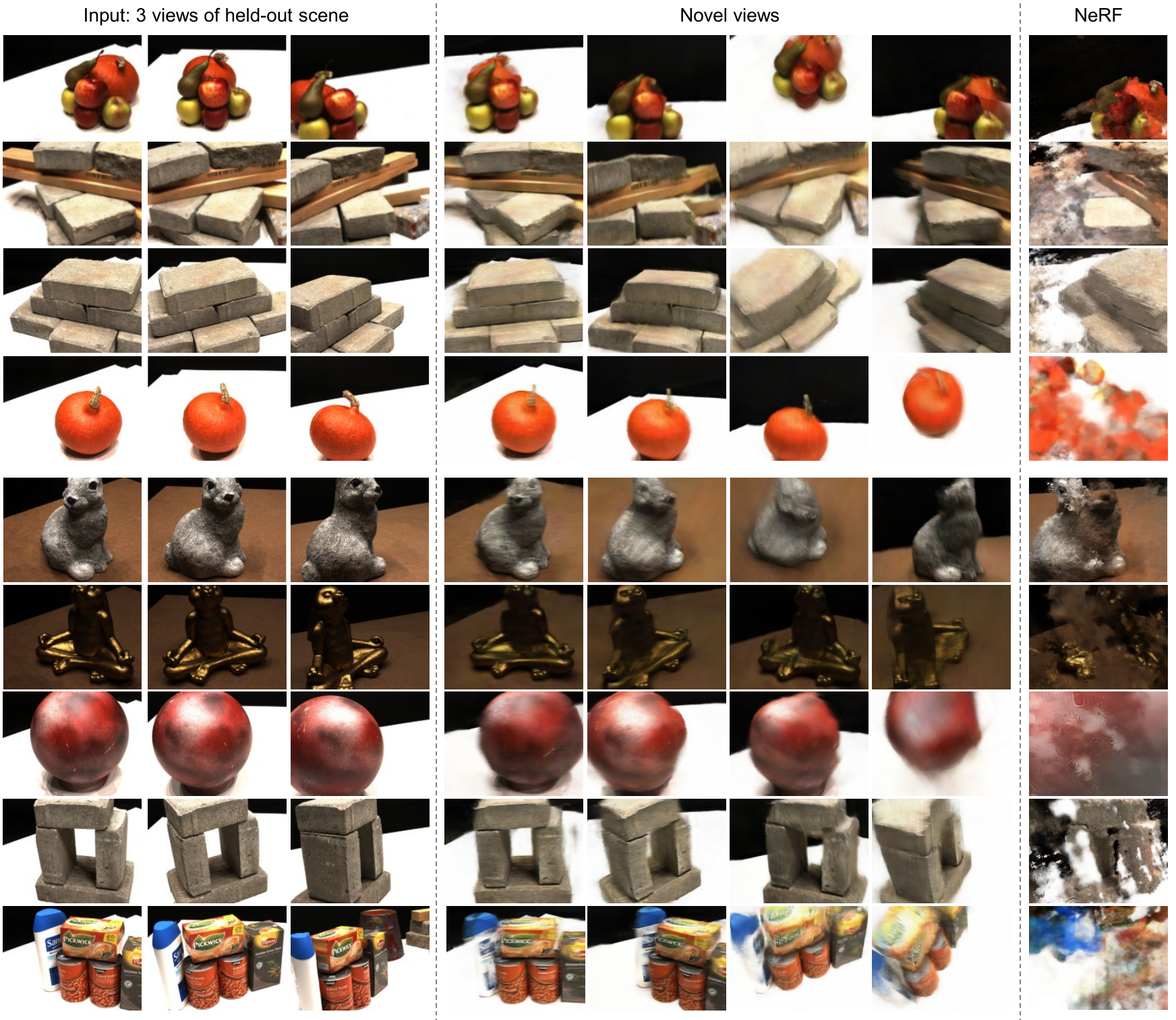}
    \caption{\textbf{Additional DTU results.} Views from the remaining 9 scenes are shown.}
    \label{fig:dtu_extra}
\end{figure*}

In Fig.~\ref{fig:dtu_extra}, we show quantitative results for each scene as well as renderings of
of all test scenes not shown in the main paper.

\begin{figure*}[t]
    \begin{center}
    \includegraphics[width=1.0\linewidth]{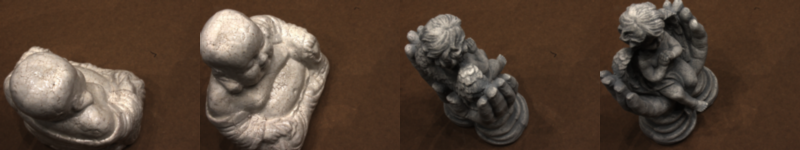}
    \end{center}
    \caption{\textbf{DTU split overlap.} The
    first and third scans (115, 119) are from the standard DTU training set from MVSNet,
    while the second and fourth (114, 118) are from the test set. 
    In our split, highly similar scenes are either all placed in the same set or 
    discarded.}
    \label{fig:dtu_bad}
    \vspace{2em}
\end{figure*}

In Table~\ref{tab:dtu_table} we provide means and standard deviations of metrics for our method and NeRF on the DTU test set, with 1, 3, 6, 9 views. The PSNR here was plotted in 
Fig.~\ref{fig:dtu_views} of the main paper
\begin{table*}[t]
    \vspace{2em}
    \begin{center}
    \setlength{\tabcolsep}{3pt}
\begin{tabular}{@{}lllllllllllllllll@{}}
\toprule
     &      & \multicolumn{3}{c}{1 View}                             &  & \multicolumn{3}{c}{3 View}                                                                                         &                      & \multicolumn{3}{c}{6 View}                                                                                         &                      & \multicolumn{3}{c}{9 View}                                                                                         \\ \cmidrule(lr){3-5} \cmidrule(lr){7-9} \cmidrule(lr){11-13} \cmidrule(l){15-17} 
     &      & PSNR & SSIM & LPIPS &  & \multicolumn{1}{c}{PSNR} & \multicolumn{1}{c}{SSIM} & \multicolumn{1}{c}{LPIPS} & \multicolumn{1}{c}{} & \multicolumn{1}{c}{PSNR} & \multicolumn{1}{c}{SSIM} & \multicolumn{1}{c}{LPIPS} & \multicolumn{1}{c}{} & \multicolumn{1}{c}{PSNR} & \multicolumn{1}{c}{SSIM} & \multicolumn{1}{c}{LPIPS} \\ \midrule
Ours & Mean & \textbf{15.55}  & \textbf{0.537}  & \textbf{0.535}     &  & \textbf{19.33}                      & \textbf{0.695}                      & \textbf{0.387}                         &                      & \textbf{20.43}                      & \textbf{0.732}                      & 0.361                                  &                      & 21.10                               & 0.758                               & 0.337                                  \\
     & SD   & 1.87            & 0.127           & 0.081              &  & 2.59                                & 0.131                               & 0.105                                  &                      & 2.66                                & 0.115                               & 0.102                                  &                      & 2.71                                & 0.102                               & 0.094                                  \\
NeRF & Mean & 8.00            & 0.286           & 0.703              &  & 9.85                                & 0.374                               & 0.622                                  &                      & 18.59                               & 0.719                               & \textbf{0.347}                         &                      & \textbf{22.14}                      & \textbf{0.820}                      & \textbf{0.262}                         \\
     & SD   & 3.20            & 0.093           & 0.055              &  & 4.69                                & 0.173                               & 0.137                                  &                      & 4.72                                & 0.177                               & 0.133                                  &                      & 4.33                                & 0.131                               & 0.109                                  \\ \bottomrule
\end{tabular}
    \end{center}
    \caption{
        \textbf{DTU aggregate metrics vs.\ NeRF}.
        Expanding on Fig.~\ref{fig:dtu_views} in the main paper,
        we compare our method to NeRF on 
        DTU test scenes quantitatively.
        Recall higher is better for PSNR and SSIM, while lower is better for LPIPS. % Removed arrows to save space
        Note that PixelNeRF is a feed-forward method,
        while a NeRF was optimized 
        for 14 hours for each scene and set of input views.
        %\ak{Add about how PixelNeRF is a feed-forward method, while NeRF here is result from 14 hour optimization. }
    }
    \label{tab:dtu_table}
    \vspace{2em}
\end{table*}

\section{Reproducibility}

\subsection{Implementation Details}

Here we describe implementation details in the interest of reproducibility.
A general remark is that due to the high compute cost,
we did not spend significant effort to tune the architecture or
training procedure, and it is possible that
variations can perform better, or that smaller models may suffice.

\paragraph{Encoder $E$}
As briefly discussed in the main paper, we use a ResNet34 backbone and extract a feature pyramid by taking the feature maps 
prior to the first pooling operation and after the first ResNet $3$ layers.
For a $H\times W$ image, the feature maps have shapes
\begin{compactenum}
    \item 
    $64\times H/2\times W/2$
    \item 
    $64\times H/4\times W/4$
    \item 
    $128\times H/8 \times W/8$
    \item 
    $256\times H/16\times W/16$
\end{compactenum}
    These are upsampled bilinearly to $H/2 \times W/2$ and concatenated into a volume of size $512 \times H/2 \times W/2$.
For a $64\times64$ image, to avoid losing too much resolution, we skip the first pooling layer, so that
the image resolutions are at $1/2, 1/2, 1/4, 1/8$ of the input rather than $1/2, 1/4, 1/8, 1/16$.
We use ImageNet pretrained weights provided through PyTorch.

\begin{figure*}[t]
    \centering
    \includegraphics[width=0.8\linewidth]{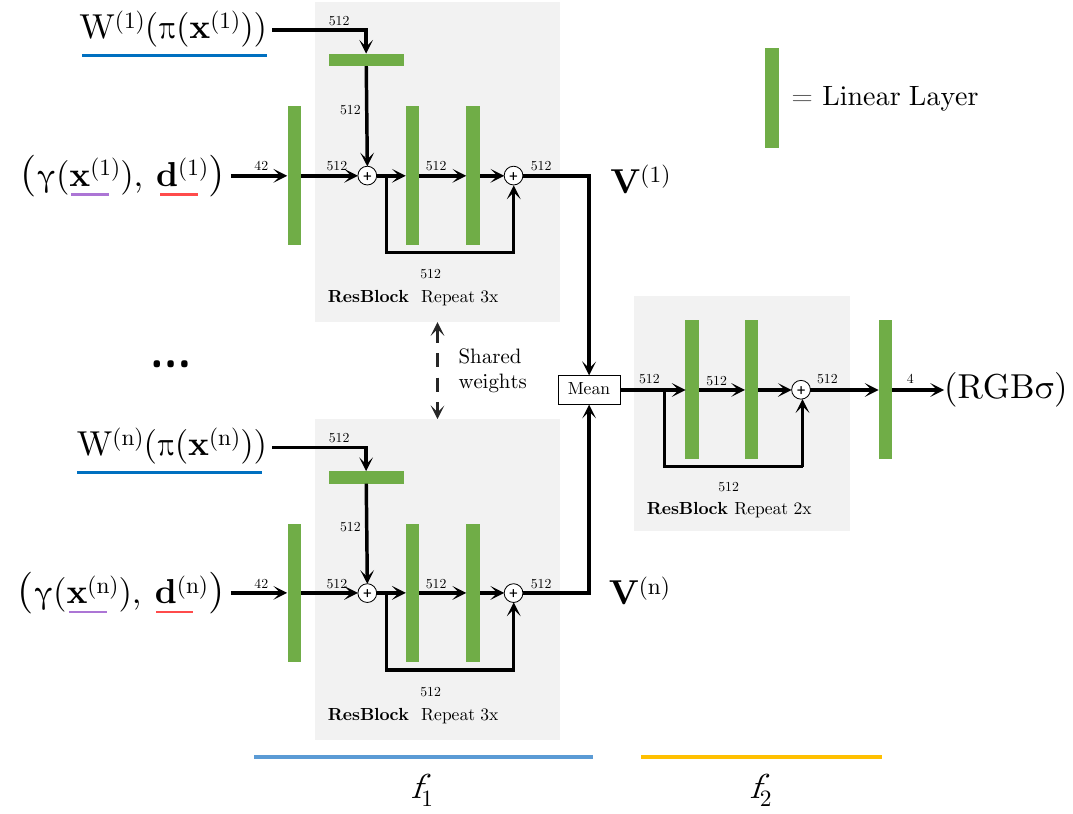}
    \caption{\textbf{Multi-view NeRF Network Architecture.}
    We use notation established in $\S$~\ref{sec:multi_cat} of the main paper, where $\gamma$ denotes a positional encoding with $6$ exponentially increasing frequencies. Each linear layer is followed by a ReLU activation.
    Note that in the single-view
    case, $f_1$ and $f_2$ can be considered a single ResNet $f = f_2 \circ f_1$.
    }
    \label{fig:nerf_net_arch}
\end{figure*}

\paragraph{NeRF network $f$}
We employ a fully-connected ResNet architecture with $5$ ResNet blocks and width $512$, similar to that in \cite{DVR}.
To enable arbitrary number of views as input, we aggregate across the source-views after block $3$ using an average-pooling operation.
This architecture is illustrated in Fig.~\ref{fig:nerf_net_arch}.
We remark that 
due to computational cost, 
we did not tune this architecture very much in practice.

\paragraph{Hierarchical volume sampling} To improve the sampling efficiency,
in practice, we also use \textit{coarse} and \textit{fine} NeRF networks $f_c, f_f$ as in the vanilla NeRF \cite{NeRF}, both of which share an identical architecture described above. Note that
the encoder $E$ is not duplicated.

More precisely, we use 64 stratified uniform and 16 importance samples,
and additionally take 16 fine samples with a normal distribution (SD 0.01) around the expected ray termination (i.e.~depth) from the coarse model, to further promote denser sampling near the surface.

\vspace{1em}
\noindent\textbf{NeRF rendering hyperparameters}
We use positional encoding $\gamma$ from NeRF for the spatial coordinates,
with exponentially increasing frequencies:
\begin{equation}
\gamma(\tf x) = 
        \begin{pmatrix}
    \sin (2^0 \omega \tf x) \\
         \cos (2^0 \omega \tf x)\\
    \sin (2^1 \omega \tf x) \\
         \cos (2^1 \omega \tf x)\\
         \vdots \\
         \sin (2^{L-1} \omega \tf x) \\
         \cos (2^{L-1} \omega \tf x)
        \end{pmatrix}
\end{equation}
Note that we do not apply the encoding to the view directions.
In all experiments, we set $L = 6$. We also
concatenate the input coordinates
along the encoding as in the NeRF implementation.
$\omega$ is a scaling factor,
set (rather arbitrarily) to $1.5$ for the
single-category, category-agnostic ShapeNet experiments as well 
as the DTU experiment,
and to $2.0$ for the multi-object experiment.
While the exponent base can be tuned, in practice we left it at $2$ as in NeRF.

The sampling bounds were set manually for 
each dataset.
They were $[1.25, 2.75]$ for ShapeNet chairs,
$[0.8, 1.8]$ for ShapeNet cars,
$[1.2, 4.0]$ for Kato et al.~\cite{NMR} renderings (category agnostic, novel category),
$[4.0, 9.0]$ for our rendered $2$-object dataset,
 and $[0.1, 5.0]$ for input.

We use a white background color in NeRF to match the ShapeNet renderings, except in the DTU setup where a black background is used.

\paragraph{Model implementation}
We implement all models using the PyTorch~\cite{PyTorch} framework.

\subsection{Experimental Details}

We first provide general details about the metrics 
and training procedure common to all experiments,
then present more specific details for each experimental setting in subsections.

\paragraph{Metric details}
We use PSNR and SSIM from the scikit-image~\cite{scikitimage} package as in SRN~\cite{SRN},
whereas LPIPS is computed with the code provided by the LPIPS authors~\cite{LPIPS} after normalizing the pixel values to the $[-1, 1]$ range.
We use the VGG network version of LPIPS following NeRF~\cite{NeRF}.

\paragraph{Training}
For all experiments, we take the learning rate to be $10^{-4}$.
We use a batch size of $4$ instances
and $128$ rays per instances.

\begin{table}[h]
    \centering
    \begin{tabular}{@{}llll@{}}
    \toprule
    \textbf{Full Name }& cabinet & display & speaker \\
    \textbf{Abbreviation} & cbnt. & disp. & spkr.\\
    \bottomrule
    \end{tabular}
    \caption{ShapeNet category name abbreviations.}
    \label{tab:name_abbreviations}
\end{table}
\vspace{-1em}

\subsubsection{Single-category ShapeNet}

We train for 400000 iterations,
which took roughly $6$ days on a single Titan RTX.
For efficiency, we sample rays 
from within a tight bounding box around the object for the first 300000 iterations, after which
we remove the bounding box to avoid background artifacts.
Further, we use $2$ input views for the first 300000 iterations
and after that,
we randomly choose to take either $1$ or $2$ views as input to encourage the model to work with either $1$ or $2$ views.

SRN's evaluation protocol is followed: in the 1-view case,
we use view $64$ as input,
and in the 2-view case, we use views $64$ and $128$.

\paragraph{Baselines}
For SRN~\cite{SRN}, we use the pretrained chair model from the public GitHub repository.
Note that SRN requires a test-time training step (latent inversion) to 
generate result images; we apply latent inversion for 170000 iterations for both the 1-view and 2-view cases for chairs.

Recall that, due to a camera sampling bug, we use an updated car dataset provided by the SRN author.
Thus, we follow instructions in the Github to train a model on the new dataset;
we train for 400000 iterations and apply latent inversion for 100000 iterations for each of the 1-view and 2-view cases.
Note the quantitative results we report are slightly lower than that in~\cite{ENR} in the single-view case, but substantially higher than in the original SRN paper, which used the bugged renderings.
For the remaining baselines, we only report numbers from the relevant papers on the same task.

\subsubsection{Category-agnostic ShapeNet}
\label{sec:detail_multicat}
We train our model for 800000 iterations on the 
entire training set, where rays are sampled from within a tight bounding box for 
the first 400000 iterations. This took about 6 days on an RTX 2080Ti.

\paragraph{Evaluation protocol}
As discussed in the main paper, we evaluate on the test split 
from~\cite{NMR} as provided by DVR~\cite{DVR}.
To ensure fairness, we sampled a random input view to encode for each object
and use this view for all baselines as well.

\paragraph{Baselines}
For DVR~\cite{DVR}, we use the pretrained 2D multiview-supervised model from the public GitHub and 
the provided rendering code (in \texttt{render.py}).
For SoftRas~\cite{SoftRas}, we similarly use the pretrained ShapeNet model from the public GitHub repo
and obtain images using their renderer library.

Since SRN~\cite{SRN} did not originally evaluate in this setting, we train a model for this category-agnostic setting using the public code.
We train for 1 million iterations and perform latent inversion for 260000 iterations,
taking about 14 days on a Titan RTX  in total.

\subsubsection{Generalization to Novel Categories}

We train our model for 680000 iterations across all instances of $3$ categories:
airplane, car, and chair. 
Rays are sampled from within a tight bounding box 
for the first 400000 iterations. This took about 5 days on a 
GTX 1080Ti.

\paragraph{Evaluation protocol}
Since there are more than 25000 objects in the 10 remaining categories,
it would be computationally prohibitive to evaluate on all of them.
For our purposes, we sample 
25\% of the objects from each category for testing, using the remaining for validation.
The protocol otherwise remains the same as in the category-agnostic model ($\S$~\ref{sec:detail_multicat}).

\paragraph{Baselines}
We train SRN and DVR as in $\S$~\ref{sec:detail_multicat}. 
For DVR~\cite{DVR}
we turned off the use of visual hull depth for sampling,
since this information was not provided for all instances of the dataset shipped with DVR.

\subsubsection{Two-object Scenes}
We generate more complex synthetic scenes consisting of two ShapeNet chairs.
We subdivide ShapeNet chairs into 2715 training instances and 1101 test instances.
We generate scenes by randomly placing instances within each split around the origin, rotated randomly about each object's z-axis, and render $128\times128$ resolution images.
Per instance, we render 20 training views sampled binned uniform on the hemisphere, and 50 testing views sampled on an Archimedean spiral, similar to the SRN protocol.

We compare with SRN~\cite{SRN} as our baseline on this task, using the publicly available code.
We note that SRN performs prediction in a canonical object-centric coordinate system, and used this version for this task.
We train a model for evaluation on this two-object dataset using one, two, and three input views.
We first train the model for 1 million iterations.
Then for each number of input views, we fix the set of input views per instance and perform latent inversion for 150,000 iterations.

\subsubsection{Sim2Real on Real Car Images}

We use car images from the Stanford Cars dataset~\cite{CarsDataset}.
PointRend~\cite{PointRend} is applied to the images
 to obtain foreground masks and bounding boxes.
 After removing the background with this mask, the image is then 
 translated and rescaled so that the center of the bounding box is at the center of the image
 and the shorter side of the bounding box
 is 1/4 of the image side length, 128. 
This normalization heuristic is motivated by the observation that the shorter side  roughly corresponds to the height or width of the car, which is a more constant quantity than the length.

For evaluation, we set the camera pose to identity
and use the same sampling strategy and bounds as at train time for the single-category cars model.

\subsubsection{Real Images on DTU}

A single model is trained on the 88 training scenes.
We use exposure level $3$ only.
Note that while there are several views per scene with 
incorrect exposure throughout the DTU dataset,
we did not remove them for training purposes.
At each training step, a random color jitter augmentation is applied equally to all views
of each object.

\paragraph{Dataset details}
While we solve a very different task 
from MVSNet~\cite{MVSNet}
which predicts depth maps from short-baseline views
and is 2.5D supervised,
we considered using
the MVSNet~\cite{MVSNet} DTU split to conform to standards for training on DTU.
However, we found that the
the split contained effective overlap across the train/val/test sets,
making it a poor benchmark of cross-scene generalization,
as shown in Fig.~\ref{fig:dtu_bad}.
For our purposes, we created a different split to avoid this issue:
we use scans 
8, 21, 30, 31, 34, 38, 40, 41, 45, 55, 63, 82, 103, 110, 114
for testing
and all other scans except
1, 2, 7, 25, 26, 27, 29, 39, 51, 54, 56, 57, 58, 73, 83, 111, 112, 113, 115, 116, 117
for training.

We downsampled all DTU images to $400 \times 300$
and adjusted the world scale of all scans by a factor or $1/300$.

\paragraph{Evaluation protocol}
We separately evaluate using 1, 3, 6, 9 informative input views and calculate image metrics with the remaining views.
Specifically, we selected views 25, 22, 28, 40, 44,48, 0, 8, 13 for input, taking a prefix of this list
when less than $9$ views are used.
We exclude the views with bad exposure (they are 3, 4, 5, 6, 7, 16, 17, 18, 19, 20, 21, 36, 37, 38, 39) for testing.

\paragraph{Baselines}
We train a total of 60 NeRFs for comparison,
one for each scene and number of input views, using the original NeRF TensorFlow code.
Each NeRF is trained for 400000 iterations
with ray batch size 128, which takes about 14 hours on an RTX Titan, to ensure convergence.

We found that NeRF did not converge in 5 cases when given 6 or 9 views,
including in the case of smurf (scan 82) shown in the video. This is possibly due to 
exposure variation in the DTU dataset.
For these scenes, we initialized the model to the trained weights for the same scene with $3$ less views
and train for about $200000$ additional iterations to get reasonable results.

\end{document}